\documentclass{article}
\usepackage{newtxtext}

    \PassOptionsToPackage{numbers, compress}{natbib}

\usepackage[preprint]{neurips_2026}

\usepackage[utf8]{inputenc} 
\usepackage[T1]{fontenc}    
\usepackage{hyperref}       
\usepackage{url}            
\usepackage{booktabs}       
\usepackage{amsfonts}       
\usepackage{nicefrac}       
\usepackage{microtype}      
\usepackage{xcolor}         

\usepackage{graphicx}
\usepackage{algorithm}
\usepackage{algpseudocode}
\usepackage{amsmath}
\usepackage{amssymb} 

\usepackage{float}

\usepackage{tabularx}
\usepackage{placeins}
\usepackage{tcolorbox}
\usepackage{enumitem}
\usepackage[table]{xcolor}
\usepackage{booktabs}
\usepackage{tabularx}
\usepackage{array}

\definecolor{PACTHeader}{RGB}{232,238,247}
\definecolor{PACTRow}{RGB}{248,250,253}
\definecolor{PACTSubRow}{RGB}{242,246,252}

\title{PACT: Proactive Asking for Continual Task Assistance in Human-Robot Collaboration}

\author{%
  Chengbo He\textsuperscript{1},
  Sheng Li\textsuperscript{1},
  Chenyang Ma\textsuperscript{2},
  Bochao Zou\textsuperscript{1},
  Li Sun\textsuperscript{1},
  Jiansheng Chen\textsuperscript{1},\\
  Junliang Xing\textsuperscript{3},
  Yuanchun Shi\textsuperscript{3},
  Huimin Ma\textsuperscript{1}\\
  \\
  \textsuperscript{1}University of Science and Technology Beijing\\
  \textsuperscript{2}University of Oxford\\
  \textsuperscript{3}Tsinghua University
}

\begin{document}

\maketitle

\begin{abstract}
Robotic assistants in long-term human-robot collaboration need to assist users under partial observations while leveraging cross-day interaction history. However, human traits and routines are often unknown at the beginning of collaboration, making passive infer-then-act assistance ineffective and inefficient. To address this challenge, we study a cross-day proactive asking setting for continual task assistance and propose \textbf{PACT} (\textbf{P}roactive \textbf{A}sking for \textbf{C}ontinual \textbf{T}ask Assistance), an ask-or-act framework that determines whether clarification should be sought before taking action. PACT leverages current observations together with accumulated interaction history to evaluate contextual sufficiency, enabling the robot to provide more reliable assistance and progressively adapt to the user over time. We implement its primary learned instantiation using reinforcement learning and evaluate alternative instantiations under the same framework. To assess such behavior, we further introduce a \textit{clarification utility} metric that quantifies the trade-off between assistance accuracy and the frequency of clarification requests. Experiments in multi-day embodied collaboration scenarios demonstrate that, compared with passive inference baselines, PACT consistently improves both assistance accuracy and clarification utility, highlighting the importance of proactive asking in continual human-robot collaboration. 
\end{abstract}

\section{Introduction}

Robotic assistants in long-term human-robot collaboration must support evolving human needs across repeated interactions and diverse collaborative activities~\citep{ayub2025continual}. Unlike short-horizon task settings, continual and open-ended collaboration requires robots to adapt to individualized human traits, preferences, and routines that may evolve over time~\citep{kekana2025review,min2024situated,chisari2025robotic}. To provide reliable and personalized assistance, robots must progressively infer collaborative task needs from partial observations and accumulated interaction history, while continuously adapting their behavior through long-term interaction~\citep{ma2025coopera}.

Recent research has begun to move beyond short, predefined tasks toward longer-horizon and more open-ended collaborative settings~\citep{matheus2025longterm,ayub2025continual,ma2025coopera}. However, most existing systems still follow a passive infer-then-act paradigm: even when the robot lacks sufficient information about the human's intent or collaborative task needs, it is forced to infer from partial observations alone and act without first seeking clarification~\citep{min2024situated,dogan2022clarifications,chisari2025robotic}. This limitation becomes particularly problematic in continual cross-day collaboration. Since human traits and routines are often unknown during early interactions, robots inevitably rely on repeated trial-and-error behaviors to gradually adapt to their users. As a result, the robot may repeatedly provide incorrect, inefficient, or even unhelpful assistance under uncertain assumptions. Such passive adaptation not only reduces collaboration efficiency, but also slows the robot's ability to personalize assistance over time. In contrast, proactive clarification can help robots identify user preferences, routines, and collaborative needs more efficiently, enabling faster adaptation and more reliable long-term assistance. Therefore, continual cross-day collaboration requires an explicit ask-or-act mechanism that determines whether the robot should act from the current context or proactively seek clarification before acting.

To address this gap, we formulate a \emph{cross-day proactive asking} setting for continual task assistance in human-robot collaboration, where a robot assists a human partner over multiple days using partial current observations and accumulated interaction history. Rather than always acting from incomplete context, the robot may proactively seek clarification when information is insufficient. The key challenge is deciding when to ask: the robot should seek clarification when uncertainty may compromise assistance quality, while avoiding unnecessary interruptions during collaboration.

\begin{figure*}[t]
  \centering
  \includegraphics[width=0.9\textwidth]{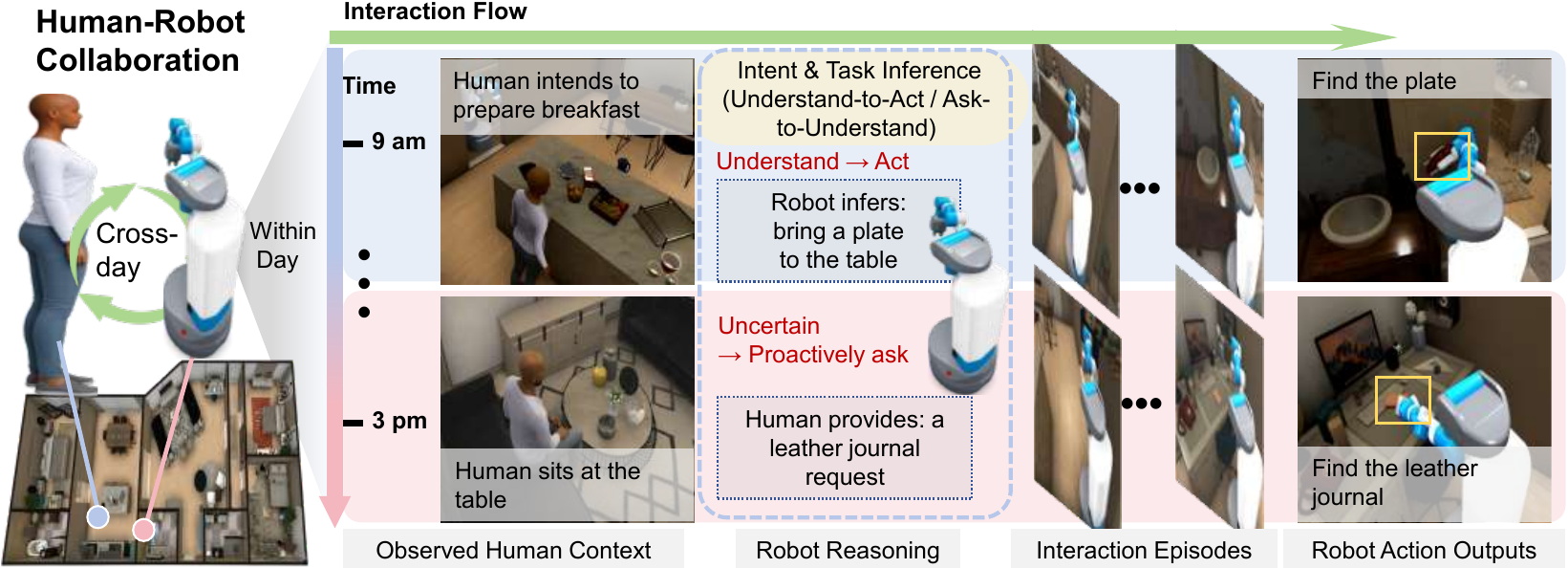}
  \caption{\textbf{Overview of PACT.} PACT studies cross-day proactive asking for continual task assistance in human-robot collaboration. The robot interacts with the human partner across days and observes multiple within-day episodes, such as morning and afternoon activities. At each time point, it combines the current human context with accumulated cross-day interaction history to infer the human's intent and task needs. Before acting, PACT performs an explicit ask-or-act decision: when the context is sufficient, the robot directly executes the inferred assistance; when the intent or task remains uncertain, it proactively asks a targeted clarification question before acting.}
  \label{fig:teaser}
\end{figure*}

Building on this setting, we propose \textbf{PACT} (\textit{Proactive Asking for Continual Task Assistance}), an ask-or-act framework for continual human-robot collaboration (Fig.~\ref{fig:teaser}). PACT formulates proactive asking as a clarification process before action execution. Specifically, the robot leverages current observations together with cross-day interaction history to assess whether the available context sufficiently specifies the human’s intent and collaborative task needs. When the context is insufficient, the robot proactively asks targeted clarification questions at either the intent level or the task level before proceeding with assistance execution. We implement the primary learned instantiation of PACT using reinforcement learning to optimize ask-or-act decisions, and evaluate alternative prompting-based, supervised, uncertainty-aware, and defer-style instantiations under the same framework. To evaluate proactive asking behavior, we further introduce a \textit{clarification utility} metric that captures the trade-off between assistance accuracy and the number of clarification questions asked, measuring whether a robot can achieve higher assistance accuracy with fewer clarification questions. We compare PACT against passive inference baselines under a multi-day collaboration protocol. Experimental results show that PACT improves assistance accuracy, highlighting the importance of cross-day proactive asking for continual human-robot assistance.

Our main contributions are as follows.
\begin{itemize}
    \item We introduce a \emph{cross-day proactive asking} setting for continual human-robot task assistance, where robots need to adapt to individualized human needs from partial observations and accumulated interaction history.

    \item We propose \textbf{PACT}, an ask-or-act framework that enables robots to decide whether clarification is necessary before assistance execution. We implement its primary learned instantiation using reinforcement learning and evaluate alternative prompting-based and training-based instantiations under the same framework.

    \item We present a reusable \textit{clarification utility} metric for evaluating proactive asking behavior, capturing the trade-off between assistance accuracy and clarification frequency.
\end{itemize}

\section{The PACT Framework}
\label{sec:method}

This section describes the PACT framework for proactive clarification in continual human-robot collaboration. We first introduce the ask-or-act formulation and the robot-observable state representation used by PACT, and summarize the full interaction procedure in Algorithm~\ref{alg:pact} (Sec.~\ref{sec:ask_or_act}). We then present Clarification Utility, which measures the trade-off between assistance accuracy and the number of clarification questions (Sec.~\ref{sec:clarification_utility}). Additional method details are provided in Appendix~\ref{app:additional_method_details}.

\begin{figure*}[t]
    \centering
    \includegraphics[width=\textwidth]{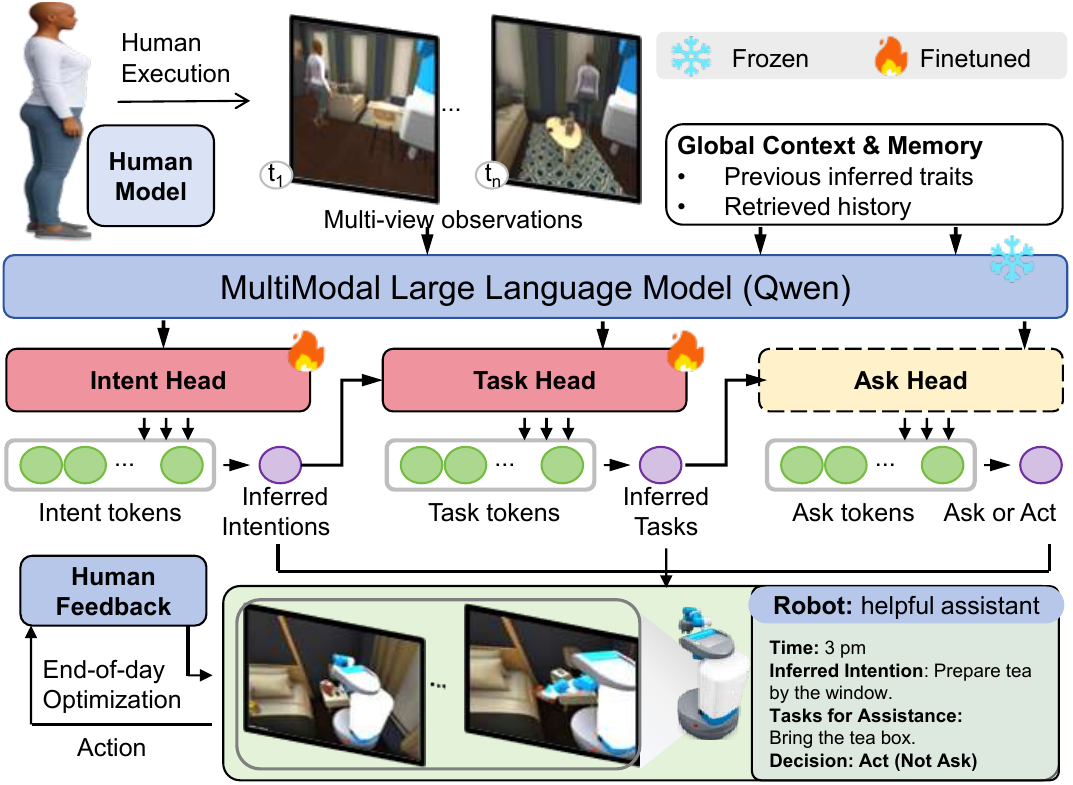}
    \caption{\textbf{PACT framework and architecture.}
    PACT uses current observations and accumulated cross-day interaction history to infer intent and task needs before acting. 
    A shared multimodal backbone provides representations to the intent, task, and ask heads; the ask head decides whether clarification is needed at the intent or task level. 
    After assistance execution, interaction outcomes and human feedback are written back into memory to support future ask-or-act decisions.}
    \label{fig:framework}
\end{figure*}

\subsection{Proactive Ask-or-Act}
\label{sec:ask_or_act}

Building on the continual task assistance setting~\citep{ma2025coopera}, PACT formulates each interaction step as an ask-or-act process under partial robot-observable context.

To instantiate this formulation, we propose the PACT framework and architecture shown in Fig.~\ref{fig:framework}. At each interaction step, the robot uses a robot-observable interaction state \(x_t=(o_t,c_t,H_{<t})\), where \(o_t\) denotes the current multimodal observations, \(c_t\) denotes the temporal context, and \(H_{<t}\) denotes the accumulated cross-day history. The observations include multi-view visual inputs of the human and surrounding environment, while the cross-day history contains previously inferred human traits and retrieved interaction records. This state provides the evidence for intent inference, task inference, and clarification estimation before acting. The multimodal large language model serves as the shared backbone of PACT. It encodes the current observations together with temporal context and cross-day history, and provides multimodal representations to the heads attached below it. The intent head uses these representations to generate a set of candidate human intents, scores them against the current interaction state, and filters them into a set of inferred intents. For each inferred intent, the task head generates a set of candidate task needs, scores them against the robot-observable interaction state \(x_t\), and filters them into inferred task needs. Together, the intent and task heads form the candidate generation, scoring, and filtering pipeline for assistance inference.

Although the intent and task heads produce inferred intents and inferred task needs, these results may still be unreliable when the current observations are partial or the cross-day history is insufficient. PACT therefore uses the ask head for clarification estimation before acting. Operationally, the ask head is applied to both intent inference and task inference. For a target \(m\in\{\mathrm{intent},\mathrm{task}\}\), it estimates a target-specific ask-or-act output:
\begin{equation}
    q_t^m =
    \pi_{\mathrm{ask}}^m(x_t,\widehat{\mathcal{Y}}_t^m,b_t),
    \qquad q_t^m\in\{0,1\},
    \label{eq:ask_policy}
\end{equation}
where \(\widehat{\mathcal{Y}}_t^m\) denotes the inferred intents or inferred task needs, and \(b_t\) is the remaining ask budget. Here, \(q_t^m=1\) means that the robot asks a clarification question about target \(m\), while \(q_t^m=0\) means that it proceeds without asking. In practice, when \(b_t=0\), we set \(q_t^m=0\). Specifically, \(q_t^{\mathrm{intent}}\) corresponds to intent clarification, and \(q_t^{\mathrm{task}}\) corresponds to task clarification. If the robot asks, it receives the corresponding human response and updates the inferred intents or inferred task needs before acting. After execution, the inferred intents, inferred task needs, ask-or-act outputs, human responses, action outcome, and feedback are appended to the cross-day history, which is used in subsequent interaction steps and days. We use \(\widetilde{\mathcal{Y}}_t^m\) to denote the final inferred intents or task needs after optional clarification, \(r_t^m\) to denote the corresponding human response, \(a_t\) to denote the executed assistance action, and \(\phi_t\) to denote the observed action outcome and feedback.

In Algorithm~\ref{alg:pact}, \(\mathcal{C}_t^m\) denotes the target-specific context: it is \(x_t\) for intent inference and \((x_t,\widetilde{\mathcal{Y}}_t^{\mathrm{intent}})\) for task inference, since the concrete task need depends on the inferred intent. \(\mathrm{Head}_m\) denotes the target-specific head that generates, scores, and filters candidates for target \(m\); \(\mathrm{Infer}_m\) updates \(\widehat{\mathcal{Y}}_t^m\) using the human response \(r_t^m\) when available, and otherwise retains \(\widehat{\mathcal{Y}}_t^m\) as the final result. Detailed candidate generation, scoring, filtering, and clarification-response construction are provided in Appendix~\ref{app:ask_or_act_details}.

\begin{algorithm}[t]
\small
\caption{Proactive ask-or-act in PACT}
\label{alg:pact}
\begin{algorithmic}[1]
\Require Robot-observable interaction state \(x_t=(o_t,c_t,H_{<t})\), ask budget \(b_t\)
\Ensure Executed assistance action \(a_t\)

\For{\(m\) in order \((\mathrm{intent}, \mathrm{task})\)}
    \If{\(m=\mathrm{intent}\)}
        \State Set target-specific context \(\mathcal{C}_t^m \gets x_t\)
    \Else
        \State Set target-specific context \(\mathcal{C}_t^m \gets (x_t,\widetilde{\mathcal{Y}}_t^{\mathrm{intent}})\)
    \EndIf

    \State \(\widehat{\mathcal{Y}}_t^m \gets \mathrm{Head}_m(\mathcal{C}_t^m)\)
    \State Estimate whether clarification is needed:
    \(q_t^m \gets \pi_{\mathrm{ask}}^m(x_t,\widehat{\mathcal{Y}}_t^m,b_t)\)
    \If{\(b_t=0\)}
        \State \(q_t^m \gets 0\)
    \EndIf

    \If{\(q_t^m=1\)}
        \State Ask a clarification question about target \(m\)
        \State Receive human response \(r_t^m\)
        \State \(b_t \gets b_t-1\)
    \Else
        \State \(r_t^m \gets \varnothing\)
    \EndIf

    \State Update inferred target after optional clarification:
    \(\widetilde{\mathcal{Y}}_t^m \gets
    \mathrm{Infer}_m(\widehat{\mathcal{Y}}_t^m,\mathcal{C}_t^m,r_t^m)\)
\EndFor

\State \(a_t \gets
\pi_{\mathrm{act}}(x_t,\widetilde{\mathcal{Y}}_t^{\mathrm{intent}},
\widetilde{\mathcal{Y}}_t^{\mathrm{task}})\)
\State Observe action outcome and feedback \(\phi_t\)
\State \(H_t \gets H_{<t}\cup
\{x_t,\widetilde{\mathcal{Y}}_t^{\mathrm{intent}},
\widetilde{\mathcal{Y}}_t^{\mathrm{task}},
q_t^{\mathrm{intent}},q_t^{\mathrm{task}},
r_t^{\mathrm{intent}},r_t^{\mathrm{task}},a_t,\phi_t\}\)
\State \Return \(a_t\)
\end{algorithmic}
\end{algorithm}

\subsection{Clarification Utility}
\label{sec:clarification_utility}

The central question in proactive asking is not only whether the robot provides correct assistance, but also whether clarification improves assistance while avoiding excessive questioning. To capture this trade-off, we introduce \textbf{Clarification Utility}, which jointly measures assistance accuracy and clarification frequency.

At interaction step \(t\), let \(A_t\in[0,1]\) denote the post-action assistance score, computed by comparing the robot's final assistance with the human-required assistance. In our evaluation, \(A_t\) is instantiated as binary assistance correctness, so \(\sum_{t=1}^{T}A_t\) corresponds to the number of correctly assisted steps. Let \(q_t\) denote the number of clarification questions asked at step \(t\), including both intent and task clarification. Given \(T\) interaction steps and \(K=\sum_{t=1}^{T}q_t\) total clarification questions, we define Clarification Utility as
\begin{equation}
    U = \frac{\sum_{t=1}^{T} A_t}{T + K}.
    \label{eq:clarification_utility}
\end{equation}

Equivalently, with
\begin{equation}
    \mathrm{Acc}=\frac{1}{T}\sum_{t=1}^{T}A_t,
    \qquad
    \mathrm{AskRate}=\frac{K}{T},
\end{equation}
Clarification Utility can be written as
\begin{equation}
    U = \frac{\mathrm{Acc}}{1+\mathrm{AskRate}}.
\end{equation}
Thus, Clarification Utility favors methods that achieve higher assistance accuracy with fewer clarification questions. Additional derivations and properties are provided in Appendix~\ref{app:clarification_utility}.

\section{Experiments}

We evaluate PACT in day-spanning human-robot collaboration scenarios involving continual task assistance. Section~\ref{sec:exp_setup} describes the method instantiation, collaboration settings, evaluation metrics, and implementation protocol. Section~\ref{sec:main_results} presents PACT instantiations, baselines, and main results under a unified ask-or-act interaction protocol across all collaboration settings. Section~\ref{sec:clarification_analysis} analyzes proactive clarification by examining how different methods balance downstream assistance quality and question frequency. 

\subsection{Experimental Setup}
\label{sec:exp_setup}

\noindent\textbf{Environment and simulated humans.}
Following the continual HRC protocol of COOPERA~\citep{ma2025coopera}, 
we instantiate dynamic household collaboration environments in Habitat 3.0~\citep{habitat3} using HSSD household scenes~\citep{hssd200}, YCB dynamic objects~\citep{ycb}, and a Fetch mobile manipulator~\citep{fetch_freight}. 
We evaluate PACT across 5 household scenes and 10 profile-conditioned simulated humans. 
The simulated humans are grounded in persistent traits, daily routines, evolving intentions, and accumulated interaction history, with embodied motions from Motion-X~\citep{lin2023motionx} and AMASS~\citep{mahmood2019amass}. 
Memory retrieval uses MiniLM-L6-v2~\citep{wang2020minilm} to retrieve relevant past intentions and tasks. 
Full environment statistics, retrieval settings, and validation results comparing simulated-human behaviors with real-user routines are provided in Appendix~\ref{app:environment_simulated_human_details}.

\noindent\textbf{Collaboration types and settings.}
We consider two collaboration types with increasing difficulty and openness. 
Collaboration Type 1 follows an open-ended Watch-and-Help-style setting, where each intent is decomposed into three pick-and-place tasks and the robot observes the first task with a textual description. 
Collaboration Type 2 is more open-ended: each intent is decomposed into five tasks involving free-form human motions and interactions with static objects, and the robot observes only the first-task video without a textual task description. 
Unlike Type 1, Type 2 is not limited to pick-and-place assistance, requiring the robot to infer a useful object or support action from under-specified human behavior. 
For both collaboration types, we evaluate four day-spanning settings: same human / same scene, same human / different scenes, different humans / same scene, and different humans / different scenes. 
These settings vary whether the human partner and the scene remain fixed or change across days, testing repeated interaction, scene adaptation, cross-human transfer, and generalization under joint human--scene variation. 
Detailed rollout structures and setting definitions are provided in Appendix~\ref{app:collaboration_settings}.

\noindent\textbf{PACT instantiation and compute.}
Figure~\ref{fig:framework} shows the model instantiation used in our experiments. 
We use Qwen3.5-27B as the shared multimodal reasoning backbone for robot-side visual-language inference, served through vLLM with an OpenAI-compatible inference interface. 
Given current observations, temporal context, candidate intentions and tasks, the remaining ask budget, and accumulated interaction history, PACT performs intent inference, task inference, and ask-or-act decision making through intent, task, and ask heads.

We implement the primary learned PACT instantiation using reinforcement learning, and also evaluate prompting-based and other training-based ASK policies for comparison. 
Prompting-based policies use the same robot-observable context and decide whether to ask through prompting, without updating ASK-specific classifier heads. 
Training-based policies implement the intent, task, and ask heads with Qwen3-8B-based classifier heads, which are updated at the end of each day using accumulated interaction data and offline supervision signals. 
The intent and task heads filter candidate intentions and task predicates, while the ask head predicts whether clarification is needed at the intent or task level.

To ensure a controlled comparison, all evaluated methods share the same observation space, collaboration settings, and evaluation pipeline. 
Classifier finetuning uses LoRA with rank 8, alpha 16, and dropout 0.2, and is trained for 5 epochs using AdamW with learning rate \(1\times10^{-5}\), weight decay 0.01, batch size 2, and gradient accumulation over 4 steps. 
All experiments are run on three NVIDIA RTX A6000 GPUs, each with 48GB memory. 
Additional online inference, day-end update, training, and compute details are provided in Appendix~\ref{app:model_training_details}.

\noindent\textbf{Evaluation protocol and metrics.}
A robot assistant interacts with a simulated human partner over multiple days. 
Each day contains 12 one-hour interaction steps from 9 a.m. to 9 p.m. 
At each step, the robot observes the current scene, the human's ongoing behavior, temporal context, and accumulated interaction history. 
It then either provides assistance directly or first asks a clarification question before acting. 
Each method is evaluated over complete multi-day interaction sequences rather than independent single-step episodes. 
This setting allows us to measure both immediate assistance quality and whether past interactions improve later assistance.

For downstream collaboration performance, we report task-level F1 and intent-level F1 across days in the main paper, with additional accuracy-based analyses in Appendix~\ref{app:additional_global_metrics}. 
Task-level F1 evaluates whether the robot identifies and supports the concrete assistance need, while intent-level F1 evaluates whether the robot identifies the human's high-level intention. 
For proactive clarification, we report ASK rate and Clarification Utility, defined in Sec.~\ref{sec:clarification_utility}. 
Together, these metrics measure final assistance quality and clarification efficiency.

\subsection{Main Results}
\label{sec:main_results}

\noindent\textbf{PACT instantiations and baselines.}
PACT is our proactive ask-or-act framework, and \textbf{ASK-RL} is our primary learned instantiation. 
It treats asking and acting as sequential decisions and optimizes the ask-or-act policy with interaction-level rewards and clarification cost. 
For comparison, we also instantiate PACT with alternative ASK policies, denoted by the prefix ``ASK-''. 
All ASK-based methods follow the same PACT interaction protocol and use the same robot-observable context; they differ only in how the ask-or-act decision is made.

The main passive baseline is \textbf{No-Ask}, which disables proactive clarification and always acts from the inferred intent and task predictions. 
This corresponds to the COOPERA-style passive infer-then-act setting~\citep{ma2025coopera}, where the robot infers assistance needs from partial observations and accumulated history without asking clarification questions.

\noindent\textbf{Alternative ASK policies.}
Prompting-based alternatives include 1) \textbf{ASK-ZeroShot:} directly prompts the backbone to decide Ask or NoAsk; 2) \textbf{ASK-FewShot:} adds labeled Ask/NoAsk examples; 3) \textbf{ASK-ToT:} uses tree-style reasoning~\citep{yao2023tree}; 4) \textbf{ASK-Proactive-CoT:} uses proactive sub-question reasoning~\citep{deng2023prompting}; and 5) \textbf{ASK-UoT:} uses uncertainty-aware reasoning~\citep{hu2024uncertainty}. 
Training-based alternatives include 6) \textbf{ASK-SFT:} learns Ask/NoAsk decisions from supervised labels~\citep{ouyang2022training}; and 7) \textbf{ASK-L2D:} treats clarification as a defer-style action~\citep{madras2018predict,mozannar2020consistent}. 
Details are provided in Appendix~\ref{app:ask_strategies}.

\noindent\textbf{Non-clarification baselines and ablations.}
Following COOPERA~\citep{ma2025coopera}, we include passive and diagnostic variants without proactive clarification. 
1) \textbf{Direct Prompting:} directly infers the intent and task need through prompting. 
2) \textbf{Direct Fine-tuning:} finetunes the model to output them without clarification. 
3) \textbf{Random:} randomly accepts candidate intents and tasks. 
4) \textbf{Oracle:} uses ground-truth intent and task information as an upper-bound reference. 
5) \textbf{Intention Agnostic:} removes explicit intent inference. 
6) \textbf{Human \& Context Agnostic:} removes persistent human-context modeling and retrieved history.

\begin{figure*}[t]
    \centering
    \includegraphics[width=\textwidth]{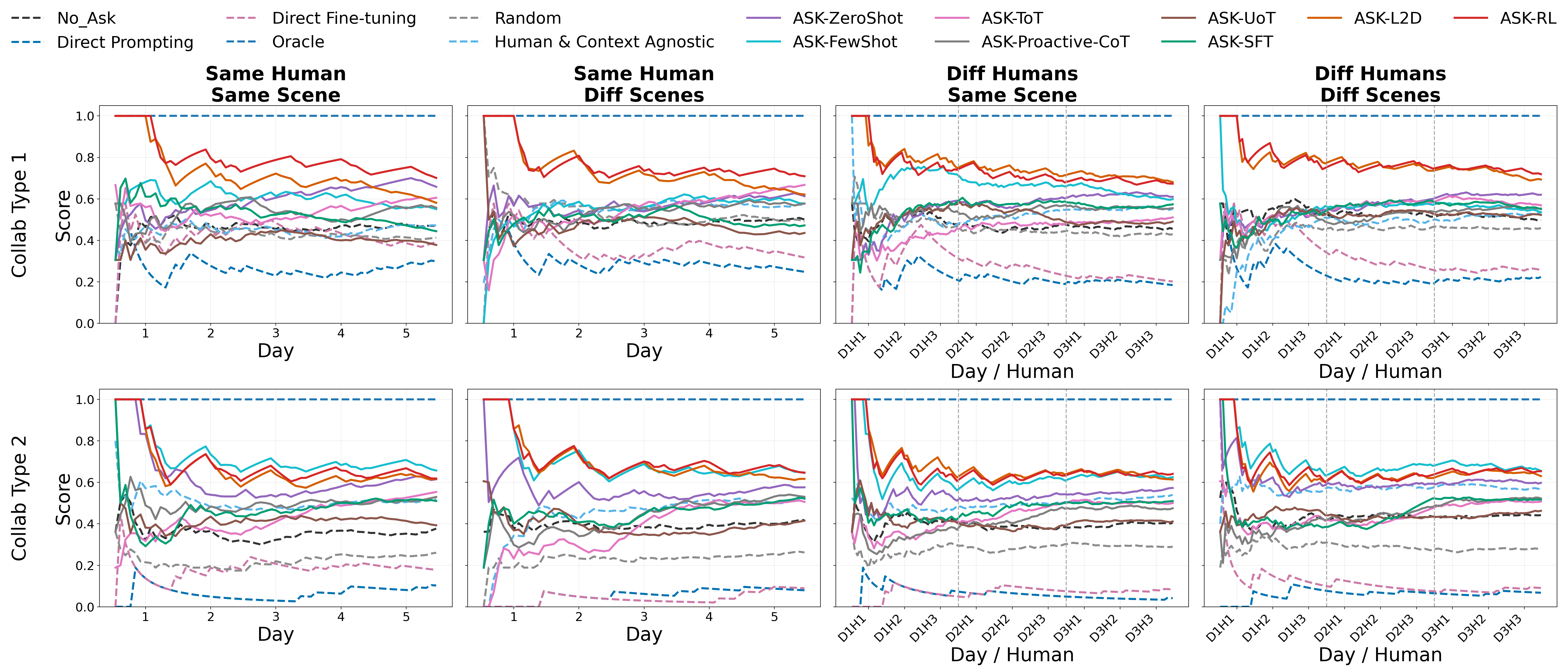}
    \caption{\textbf{Intent-level performance under day-spanning human-robot collaboration.}
    Intent-level F1 evaluates whether the assistant infers the human partner's high-level intent. 
    Rows correspond to the two collaboration types, and columns correspond to the four human-scene settings. 
    Solid curves denote ASK-based PACT instantiations, while dashed curves denote non-clarification variants.}
    \label{fig:main_intent_f1}
\end{figure*}

\begin{figure*}[t]
    \centering
    \includegraphics[width=\textwidth]{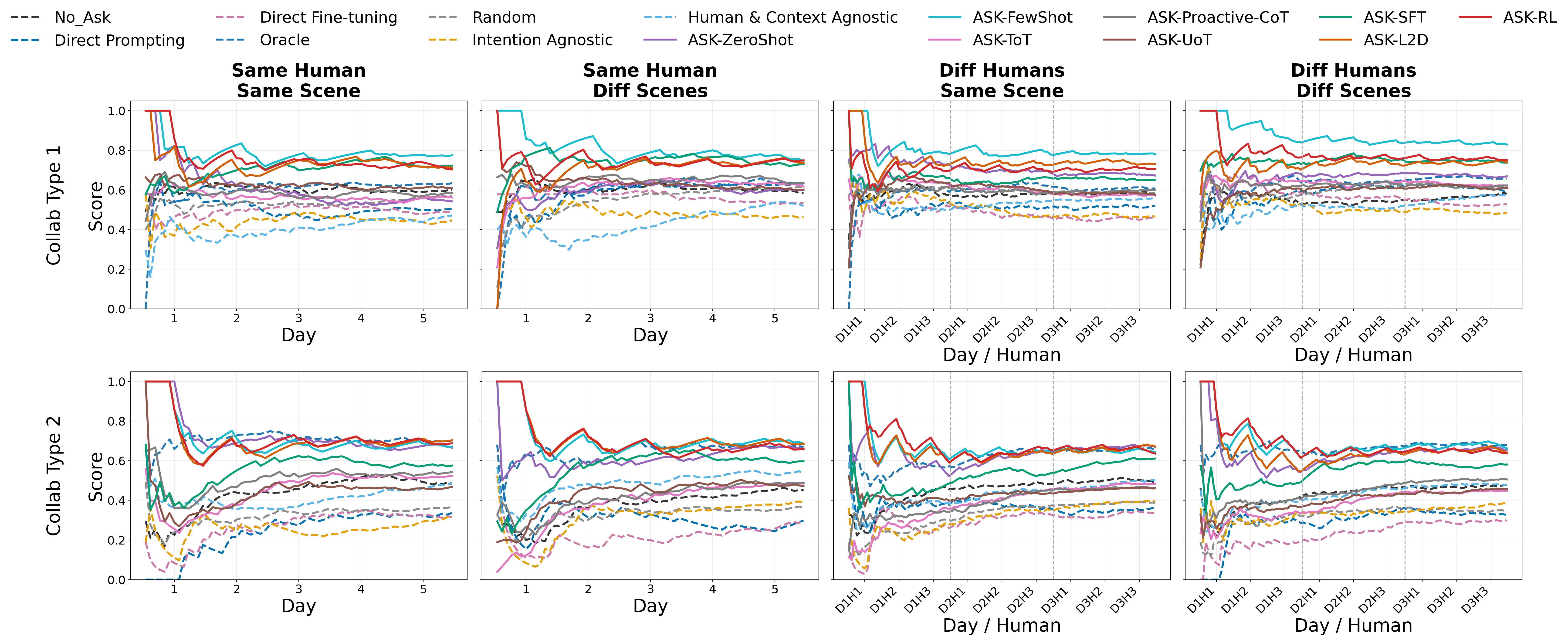}
    \caption{\textbf{Task-level performance under day-spanning human-robot collaboration.}
    Task-level F1 evaluates whether the assistant infers the concrete assistance need before acting. 
    Rows correspond to the two collaboration types, and columns correspond to the four human-scene settings. 
    Solid curves denote ASK-based PACT instantiations, while dashed curves denote non-clarification variants.}
    \label{fig:main_task_f1}
\end{figure*}

Figures~\ref{fig:main_intent_f1} and~\ref{fig:main_task_f1} report intent-level and task-level F1 across four collaboration settings and two collaboration types. 
Appendix~\ref{app:additional_global_metrics} provides corresponding accuracy-based results in Figs.~\ref{fig:app_main_intent_accuracy} and~\ref{fig:app_main_task_accuracy}, which show consistent trends.

At the intent level, Fig.~\ref{fig:main_intent_f1} shows that ASK-based PACT instantiations are generally more stable than non-clarification variants, especially when the human partner, the scene, or both change across days. 
This suggests that proactive clarification helps recover missing information when accumulated history is insufficient for the current human-scene configuration. 
The weaker performance of \textbf{Intention Agnostic} and \textbf{Human \& Context Agnostic} further indicates that both intent modeling and persistent human-context tracking are important for continual assistance.

At the task level, Fig.~\ref{fig:main_task_f1} shows that ASK-based methods generally outperform non-clarification variants, with clearer gains in Collab Type 2, where passive inference remains lower throughout the rollout. 
Among ASK-based methods, \textbf{ASK-RL} and \textbf{ASK-L2D} provide consistently strong performance across both intent-level and task-level evaluation. 
This suggests that learned ask-or-act policies can improve assistance not only by resolving concrete task uncertainty, but also by maintaining more reliable high-level intent understanding.

However, strong task-level performance does not always imply efficient clarification. 
For example, \textbf{ASK-FewShot} achieves strong task-level F1 in several settings, but its weaker intent-level behavior suggests that it often improves concrete task prediction by asking more task-level clarification questions, rather than by improving the full intent-to-task reasoning process. 
In contrast, \textbf{ASK-RL} and \textbf{ASK-L2D} maintain stronger performance across both intent-level and task-level evaluation. 
This motivates a clarification-efficiency analysis: whether performance gains come from selective clarification or simply from asking more often.

\begin{figure*}[t]
    \centering
    \includegraphics[width=\textwidth]{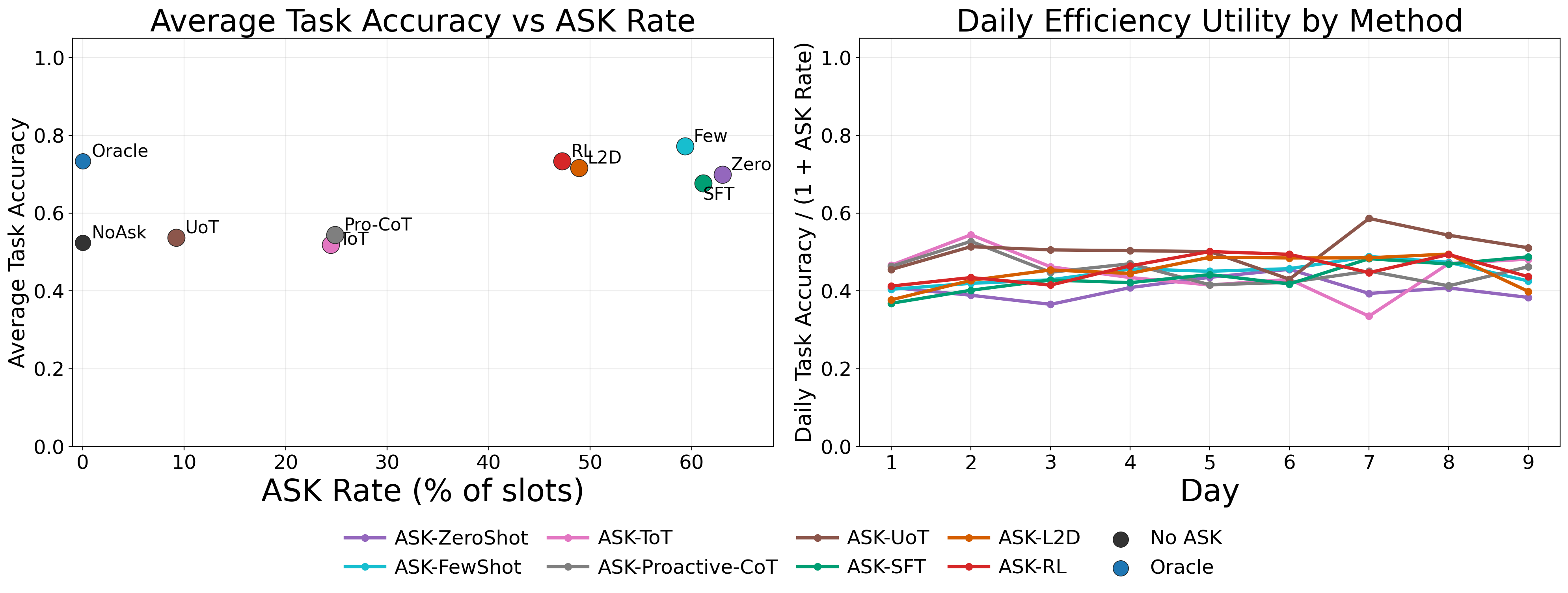}
    \caption{\textbf{Clarification efficiency in continual human-robot assistance.}
    This figure complements the intent- and task-level results in Figs.~\ref{fig:main_intent_f1} and~\ref{fig:main_task_f1} by comparing downstream task accuracy with clarification frequency. 
    Left: average task accuracy versus ASK rate across PACT instantiations and baselines. 
    Right: daily Clarification Utility, computed as task accuracy divided by \(1+\)ASK rate. 
    Higher utility indicates stronger task assistance with fewer clarification questions.}
    \label{fig:efficiency_summary}
\end{figure*}

\subsection{Proactive Clarification Analysis}
\label{sec:clarification_analysis}

Figure~\ref{fig:efficiency_summary} provides two aggregate views of clarification efficiency: average task accuracy versus ASK rate, and daily Clarification Utility. 
ASK rate measures the percentage of interaction slots with clarification, while Clarification Utility captures the trade-off between assistance accuracy and question frequency. 
The left panel shows that higher ASK rate does not necessarily imply better efficiency: ASK-FewShot achieves high task accuracy but asks most frequently, and ASK-ZeroShot asks often without proportional gains. 
In contrast, \textbf{ASK-RL}, our primary learned instantiation, and ASK-L2D achieve strong task accuracy with lower ASK rates, suggesting a better accuracy--question trade-off.

The right panel provides a complementary view of daily utility. 
ASK-UoT obtains strong utility on several later days, suggesting that uncertainty-aware reasoning is useful under cross-human variation. 
Because UoT is designed for proactive information seeking, it helps the model inspect what remains uncertain before deciding whether to ask. 
However, its lower average task accuracy in the left panel shows that high daily utility does not necessarily imply the strongest overall assistance performance. 
Thus, ASK-UoT provides a useful uncertainty signal, while ASK-RL and ASK-L2D offer a stronger overall balance between task accuracy and clarification frequency.

These results show that proactive clarification is not simply about asking more questions. 
Effective methods should ask when the current context is insufficient for reliable action, rather than relying on frequent clarification. 
Appendix~\ref{app:clarification_by_setting} provides per-setting analyses, including task accuracy--ASK rate trade-offs and Clarification Utility curves for each collaboration type and human-scene setting.

\section{Discussion and Limitations}
\label{sec:discussion}

Proactive asking in continual assistance is better viewed as selective information acquisition than as question generation. Figs.~\ref{fig:main_intent_f1} and~\ref{fig:main_task_f1} show that ASK-based PACT variants generally improve assistance over passive non-clarification baselines, while Fig.~\ref{fig:efficiency_summary} shows that a higher ASK rate does not necessarily lead to better clarification efficiency. This reveals two failure modes: under-asking, where the assistant acts under unresolved uncertainty, and over-asking, where accuracy improves mainly through excessive clarification. Thus, the key issue is not whether the robot can ask, but whether it can identify when clarification is useful for the current interaction.

The additional analyses explain this trade-off from three diagnostic perspectives. 
\textbf{ASK usage} studies where clarification is allocated, distinguishing whether a method mainly asks about intent uncertainty, task uncertainty, or both. 
\textbf{ASK policy quality} examines whether the ASK rate is adjusted over time as interaction history accumulates. 
\textbf{ASK impact} measures whether clarification leads to downstream gains in semantic similarity, intent F1, task F1, and Clarification Utility. 
These diagnostics, detailed in Appendix~\ref{app:per_setting_ask_usage}, Appendix~\ref{app:per_setting_ask_policy_quality}, and Appendix~\ref{app:per_setting_ask_impact}, show that effective proactive assistance depends on useful and selective clarification, not on asking more questions.

These analyses also help explain the difference between prompting-based and learned ASK policies. 
Prompting-based methods may either over-ask through frequent task-level clarification or under-ask when they are too conservative under uncertainty. By contrast, learned ask-or-act policies such as ASK-RL and ASK-L2D often show a more adaptive pattern: they ask in uncertain stages and rely more on accumulated context when previous interactions become informative. This behavior is important in cross-day collaboration, where history can be helpful in stable same-human settings but less reliable when the human partner, scene, or both change across days.

One limitation is that PACT is evaluated in controlled multi-day, single-user assistance settings. This design makes the comparison clear and fair across different human-scene conditions, but it does not cover all possible long-term collaboration settings, such as multi-user assistance or more open-ended daily routines. Extending PACT to these broader settings is a natural direction for future work.

Clarification Utility also uses the same cost for each clarification question. This makes the metric simple and easy to compare across methods, but it does not model finer differences between intent-level and task-level questions. Future work could add different question costs while keeping the metric simple.

\section{Conclusion}
\label{sec:conclusion}

We presented PACT, a framework for proactive asking in continual human-robot task assistance. PACT studies cross-day collaboration where the robot must assist from partial current observations and accumulated interaction history, and formulates each interaction as an ask-or-act decision before assistance execution. 
We evaluated PACT under a validated profile-conditioned simulated-human protocol, which supports controlled multi-day comparison across different human-scene settings. We implemented its primary learned instantiation with reinforcement learning and evaluated prompting-based, uncertainty-aware, supervised, and defer-style alternatives under the same protocol. 
To measure clarification behavior, we introduced Clarification Utility, which captures the trade-off between assistance accuracy and clarification frequency.

Experiments across multi-day human-scene settings show that ASK-based PACT variants generally improve intent and task inference over passive non-clarification baselines. 
The clarification analysis further shows that effective proactive assistance does not come from asking more questions, but from asking selectively when the current context is insufficient for reliable action. 
Overall, these results highlight proactive asking as an important mechanism for reliable and efficient continual human-robot assistance.

\bibliographystyle{plainnat}
\bibliography{references}

@String{Computing = "Computing" }

@String{Computer = "{IEEE} Computer" }

@article{ma2025coopera,
  title={Coopera: Continual open-ended human-robot assistance},
  author={Ma, Chenyang and Lu, Kai and Desai, Ruta and Puig, Xavier and Markham, Andrew and Trigoni, Niki},
  journal={arXiv preprint arXiv:2510.23495},
  year={2025}
}

@article{kekana2025review,
  title={A Review of Human Intention Recognition Frameworks in Industrial Collaborative Robotics},
  author={Kekana, Mokone and Du, Shengzhi and Steyn, Nico and Benali, Abderraouf and Djerroud, Halim},
  journal={Robotics},
  volume={14},
  number={12},
  pages={174},
  year={2025},
  publisher={MDPI}
}

@inproceedings{chisari2025robotic,
  title={Robotic task ambiguity resolution via natural language interaction},
  author={Chisari, Eugenio and Von Hartz, Jan Ole and Despinoy, Fabien and Valada, Abhinav},
  booktitle={2025 IEEE/RSJ International Conference on Intelligent Robots and Systems (IROS)},
  pages={14821--14827},
  year={2025},
  organization={IEEE}
}

@article{matheus2025longterm,
  author  = {Matheus, Kayla and Ramnauth, Rebecca and Scassellati, Brian and Salomons, Nicole},
  title   = {Long-Term Interactions with Social Robots: Trends, Insights, and Recommendations},
  journal = {ACM Transactions on Human-Robot Interaction},
  volume  = {14},
  number  = {3},
  pages   = {1--42},
  year    = {2025},
  doi     = {10.1145/3729539}
}

@article{ayub2025continual,
  author  = {Ayub, Ali and De Francesco, Zachary and Holthaus, Patrick and Nehaniv, Chrystopher L. and Dautenhahn, Kerstin},
  title   = {Continual Learning Through Human-Robot Interaction: Human Perceptions of a Continual Learning Robot in Repeated Interactions},
  journal = {International Journal of Social Robotics},
  volume  = {17},
  number  = {2},
  pages   = {277--296},
  year    = {2025},
  doi     = {10.1007/s12369-025-01214-9}
}

@inproceedings{min2024situated,
  author    = {Min, So Yeon and Puig, Xavi and Chaplot, Devendra Singh and Yang, Tsung-Yen and Rai, Akshara and Parashar, Priyam and Salakhutdinov, Ruslan and Bisk, Yonatan and Mottaghi, Roozbeh},
  title     = {Situated Instruction Following},
  booktitle = {Computer Vision -- ECCV 2024},
  pages     = {202--228},
  year      = {2024},
  doi       = {10.1007/978-3-031-73030-6_12}
}

@inproceedings{dogan2022clarifications,
  author    = {Dogan, Fethiye Irmak and Saffari, Amir Hossein and Hough, Julian and Leite, Iolanda},
  title     = {Asking Follow-Up Clarifications to Resolve Ambiguities in Human-Robot Conversation},
  booktitle = {Proceedings of the ACM/IEEE International Conference on Human-Robot Interaction (HRI)},
  pages     = {461--469},
  year      = {2022},
  doi       = {10.1109/HRI53351.2022.9889368}
}

@article{vaccaro2024combinations,
  title={When combinations of humans and AI are useful: A systematic review and meta-analysis},
  author={Vaccaro, Michelle and Almaatouq, Abdullah and Malone, Thomas},
  journal={Nature Human Behaviour},
  volume={8},
  number={12},
  pages={2293--2303},
  year={2024},
  publisher={Nature Publishing Group UK London}
}

@inproceedings{dhillon2024shaping,
  title={Shaping human-AI collaboration: Varied scaffolding levels in co-writing with language models},
  author={Dhillon, Paramveer S and Molaei, Somayeh and Li, Jiaqi and Golub, Maximilian and Zheng, Shaochun and Robert, Lionel Peter},
  booktitle={Proceedings of the 2024 CHI conference on human factors in computing systems},
  pages={1--18},
  year={2024}
}

@article{hwang202580,
  title={'It was 80\% me, 20\% AI': Seeking Authenticity in Co-Writing with Large Language Models},
  author={Hwang, Angel Hsing-Chi and Liao, Q Vera and Blodgett, Su Lin and Olteanu, Alexandra and Trischler, Adam},
  journal={Proceedings of the ACM on Human-Computer Interaction},
  volume={9},
  number={2},
  pages={1--41},
  year={2025},
  publisher={ACM New York, NY, USA}
}

@article{wan2024felt,
  title={" It Felt Like Having a Second Mind": Investigating Human-AI Co-creativity in Prewriting with Large Language Models},
  author={Wan, Qian and Hu, Siying and Zhang, Yu and Wang, Piaohong and Wen, Bo and Lu, Zhicong},
  journal={Proceedings of the ACM on human-computer interaction},
  volume={8},
  number={CSCW1},
  pages={1--26},
  year={2024},
  publisher={ACM New York, NY, USA}
}

@inproceedings{skantze2025applying,
  title={Applying general turn-taking models to conversational human-robot interaction. In 2025 ACM},
  author={Skantze, G and Irfan, B},
  booktitle={IEEE international conference on human-robot interaction},
  year={2025}
}

@article{cao2026reframing,
  title={Reframing Conversational Design in HRI: Deliberate Design with AI Scaffolds},
  author={Cao, Shiye and Moon, Jiwon and Xu, Yifan and Liu, Anqi and Huang, Chien-Ming},
  journal={arXiv preprint arXiv:2601.12084},
  year={2026}
}

@article{gessler2025overcookedv2,
  title={Overcookedv2: Rethinking overcooked for zero-shot coordination},
  author={Gessler, Tobias and Dizdarevic, Tin and Calinescu, Ani and Ellis, Benjamin and Lupu, Andrei and Foerster, Jakob Nicolaus},
  journal={arXiv preprint arXiv:2503.17821},
  year={2025}
}

@article{liang2024learning,
  title={Learning to cooperate with humans using generative agents},
  author={Liang, Yancheng and Chen, Daphne and Gupta, Abhishek and Du, Simon S and Jaques, Natasha},
  journal={Advances in Neural Information Processing Systems},
  volume={37},
  pages={60061--60087},
  year={2024}
}

@inproceedings{zhu2025multiagentbench,
  title={Multiagentbench: Evaluating the collaboration and competition of llm agents},
  author={Zhu, Kunlun and Du, Hongyi and Hong, Zhaochen and Yang, Xiaocheng and Guo, Shuyi and Wang, Daisy Zhe and Wang, Zhenhailong and Qian, Cheng and Tang, Robert and Ji, Heng and others},
  booktitle={Proceedings of the 63rd Annual Meeting of the Association for Computational Linguistics (Volume 1: Long Papers)},
  pages={8580--8622},
  year={2025}
}

@inproceedings{testoni2024asking,
  title={Asking the right question at the right time: Human and model uncertainty guidance to ask clarification questions},
  author={Testoni, Alberto and Fern{\'a}ndez, Raquel},
  booktitle={Proceedings of the 18th Conference of the European Chapter of the Association for Computational Linguistics (Volume 1: Long Papers)},
  pages={258--275},
  year={2024}
}

@article{chang2024partnr,
  title={Partnr: A benchmark for planning and reasoning in embodied multi-agent tasks},
  author={Chang, Matthew and Chhablani, Gunjan and Clegg, Alexander and Cote, Mikael Dallaire and Desai, Ruta and Hlavac, Michal and Karashchuk, Vladimir and Krantz, Jacob and Mottaghi, Roozbeh and Parashar, Priyam and others},
  journal={arXiv preprint arXiv:2411.00081},
  year={2024}
}

@inproceedings{yang2025embodiedbench,
  title={EmbodiedBench: Comprehensive Benchmarking Multi-modal Large Language Models for Vision-Driven Embodied Agents},
  author={Yang, Rui and Chen, Hanyang and Zhang, Junyu and Zhao, Mark and Qian, Cheng and Wang, Kangrui and Wang, Qineng and Koripella, Teja Venkat and Movahedi, Marziyeh and Li, Manling and others},
  booktitle={International Conference on Machine Learning},
  pages={70576--70631},
  year={2025},
  organization={PMLR}
}

@article{ramrakhya2025grounding,
  title={Grounding multimodal llms to embodied agents that ask for help with reinforcement learning},
  author={Ramrakhya, Ram and Chang, Matthew and Puig, Xavier and Desai, Ruta and Kira, Zsolt and Mottaghi, Roozbeh},
  journal={arXiv preprint arXiv:2504.00907},
  year={2025}
}

@inproceedings{lu2025thinkbot,
  title     = {ThinkBot: Embodied Instruction Following with Thought Chain Reasoning},
  author    = {Lu, Guanxing and Wang, Ziwei and Liu, Changliu and Lu, Jiwen and Tang, Yansong},
  booktitle = {International Conference on Learning Representations (ICLR)},
  year      = {2025}
}

@article{monwilliams2025ellmer,
  title   = {Embodied large language models enable robots to complete complex tasks in unpredictable environments},
  author  = {Mon-Williams, Ruaridh and Li, Gen and Long, Ran and Du, Wenqian and Lucas, Christopher G.},
  journal = {Nature Machine Intelligence},
  volume  = {7},
  pages   = {592--601},
  year    = {2025}
}

@inproceedings{wang2025affordbot,
  title     = {AffordBot: 3D Fine-grained Embodied Reasoning via Multimodal Large Language Models},
  author    = {Wang, Xinyi and Yang, Xun and Xu, Yanlong and Wu, Yuchen and Li, Zhen and Zhao, Na},
  booktitle = {Advances in Neural Information Processing Systems (NeurIPS)},
  year      = {2025}
}

@article{ying2024siftom,
  title   = {SIFToM: Robust Spoken Instruction Following through Theory of Mind},
  author  = {Ying, Lance and Liu, Jason Xinyu and Aarya, Shivam and Fang, Yizirui and Tellex, Stefanie and Tenenbaum, Joshua B. and Shu, Tianmin},
  journal = {arXiv preprint arXiv:2409.10849},
  year    = {2024}
}

@inproceedings{wan2025infer,
  title     = {Infer Human's Intentions Before Following Natural Language Instructions},
  author    = {Wan, Yanming and Wu, Yue and Wang, Yiping and Mao, Jiayuan and Jaques, Natasha},
  booktitle = {Proceedings of the AAAI Conference on Artificial Intelligence},
  volume    = {39},
  pages     = {25474--25482},
  year      = {2025}
}

@inproceedings{taioli2025collaborative,
  title={Collaborative instance object navigation: Leveraging uncertainty-awareness to minimize human-agent dialogues},
  author={Taioli, Francesco and Zorzi, Edoardo and Franchi, Gianni and Castellini, Alberto and Farinelli, Alessandro and Cristani, Marco and Wang, Yiming},
  booktitle={Proceedings of the IEEE/CVF International Conference on Computer Vision},
  pages={18781--18792},
  year={2025}
}

@inproceedings{singh2022ask4help,
  title     = {Ask4Help: Learning to Leverage an Expert for Embodied Tasks},
  author    = {Singh, Kunal Pratap and Weihs, Luca and Herrasti, Alvaro and Choi, Jonghyun and Kembhavi, Aniruddha and Mottaghi, Roozbeh},
  booktitle = {Advances in Neural Information Processing Systems},
  year      = {2022}
}

@article{yu2025mixed,
  title={Mixed-Initiative Dialog for Human-Robot Collaborative Manipulation},
  author={Yu, Albert and Li, Chengshu and Macesanu, Luca and Balaji, Arnav and Ray, Ruchira and Mooney, Raymond and Mart{\'\i}n-Mart{\'\i}n, Roberto},
  journal={arXiv preprint arXiv:2508.05535},
  year={2025}
}

@inproceedings{habitat3,
  title     = {Habitat 3.0: A Co-Habitat for Humans, Avatars, and Robots},
  author    = {Puig, Xavier and Undersander, Eric and Szot, Andrew and Cote, Mikael Dallaire and Yang, Tsung-Yen and Partsey, Ruslan and Desai, Ruta and Clegg, Alexander and Hlavac, Michal and Min, So Yeon and Vondrus, Vladimir and Gervet, Theophile and Berges, Vincent-Pierre and Turner, John and Maksymets, Oleksandr and Kira, Zsolt and Kalakrishnan, Mrinal and Malik, Jitendra and Chaplot, Devendra Singh and Jain, Unnat and Batra, Dhruv and Rai, Akshara and Mottaghi, Roozbeh},
  booktitle = {International Conference on Learning Representations},
  year      = {2024}
}

@inproceedings{hssd200,
  title     = {Habitat Synthetic Scenes Dataset (HSSD-200): An Analysis of 3D Scene Scale and Realism Tradeoffs for ObjectGoal Navigation},
  author    = {Khanna, Mukul and Mao, Yongsen and Jiang, Hanxiao and Haresh, Sanjay and Shacklett, Brennan and Batra, Dhruv and Clegg, Alexander and Undersander, Eric and Chang, Angel X. and Savva, Manolis},
  booktitle = {Proceedings of the IEEE/CVF Conference on Computer Vision and Pattern Recognition},
  pages     = {16384--16393},
  year      = {2024}
}

@inproceedings{ycb,
  title     = {The YCB Object and Model Set: Towards Common Benchmarks for Manipulation Research},
  author    = {Calli, Berk and Singh, Arjun and Walsman, Aaron and Srinivasa, Siddhartha and Abbeel, Pieter and Dollar, Aaron M.},
  booktitle = {Proceedings of the IEEE International Conference on Advanced Robotics},
  pages     = {510--517},
  year      = {2015}
}

@inproceedings{fetch_freight,
  title     = {Fetch \& Freight: Standard Platforms for Service Robot Applications},
  author    = {Wise, Melonee and Ferguson, Michael and King, Derek and Diehr, Eric and Dymesich, David},
  booktitle = {Workshop on Autonomous Mobile Service Robots, held at the International Joint Conference on Artificial Intelligence},
  pages     = {1--6},
  year      = {2016}
}

@inproceedings{lin2023motionx,
  title     = {Motion-X: A Large-scale 3D Expressive Whole-body Human Motion Dataset},
  author    = {Lin, Jing and Zeng, Ailing and Lu, Shunlin and Cai, Yuanhao and Zhang, Ruimao and Wang, Haoqian and Zhang, Lei},
  booktitle = {Advances in Neural Information Processing Systems},
  year      = {2023}
}

@inproceedings{mahmood2019amass,
  title     = {AMASS: Archive of Motion Capture as Surface Shapes},
  author    = {Mahmood, Naureen and Ghorbani, Nima and Troje, Nikolaus F. and Pons-Moll, Gerard and Black, Michael J.},
  booktitle = {Proceedings of the IEEE/CVF International Conference on Computer Vision},
  pages     = {5442--5451},
  year      = {2019}
}

@inproceedings{wang2020minilm,
  title     = {MiniLM: Deep Self-Attention Distillation for Task-Agnostic Compression of Pre-Trained Transformers},
  author    = {Wang, Wenhui and Wei, Furu and Dong, Li and Bao, Hangbo and Yang, Nan and Zhou, Ming},
  booktitle = {Advances in Neural Information Processing Systems},
  year      = {2020}
}

@ArtifactSoftware{R,
    title = {R: A Language and Environment for Statistical Computing},
    author = {{R Core Team}},
    organization = {R Foundation for Statistical Computing},
    address = {Vienna, Austria},
    year = {2019},
    url = {https://www.R-project.org/},
}

@inproceedings{Mozannar2020Consistent,
  author = {Hussein Mozannar and David Sontag},
  title = {Consistent Estimators for Learning to Defer to an Expert},
  booktitle = {International Conference on Machine Learning},
  pages = {7076--7087},
  year = {2020}
}

@inproceedings{xiong2025deliberate,
  title = {Deliberate Reasoning in Language Models as Structure-Aware Planning with an Accurate World Model},
  author = {Xiong, Siheng and Payani, Ali and Yang, Yuan and Fekri, Faramarz},
  booktitle = {Proceedings of the 63rd Annual Meeting of the Association for Computational Linguistics (Volume 1: Long Papers)},
  pages = {31900--31931},
  year = {2025},
  month = jul,
  publisher = {Association for Computational Linguistics},
  doi = {10.18653/v1/2025.acl-long.1540},
  url = {https://aclanthology.org/2025.acl-long.1540/}
}

@inproceedings{xiong2026enhancing,
  title = {Enhancing Language Model Reasoning with Structured Multi-Level Modeling},
  author = {Xiong, Siheng and Payani, Ali and Fekri, Faramarz},
  booktitle = {The Fourteenth International Conference on Learning Representations},
  year = {2026},
  url = {https://openreview.net/forum?id=PlkzZhqBCd}
}

@article{xiong2026scaling,
  title = {Scaling Search-Augmented LLM Reasoning via Adaptive Information Control},
  author = {Xiong, Siheng and Gungordu, Oguzhan and Johnson, Blair and Kerce, James C. and Fekri, Faramarz},
  journal = {arXiv preprint arXiv:2602.01672},
  year = {2026},
  url = {https://arxiv.org/abs/2602.01672}
}

@inproceedings{yao2023tree,
  title={Tree of Thoughts: Deliberate Problem Solving with Large Language Models},
  author={Yao, Shunyu and Yu, Dian and Zhao, Jeffrey and Shafran, Izhak and Griffiths, Thomas L. and Cao, Yuan and Narasimhan, Karthik},
  booktitle={Advances in Neural Information Processing Systems},
  year={2023}
}

@inproceedings{deng2023prompting,
  title={Prompting and Evaluating Large Language Models for Proactive Dialogues: Clarification, Target-guided, and Non-collaboration},
  author={Deng, Yang and Liao, Lizi and Chen, Liang and Wang, Hongru and Lei, Wenqiang and Chua, Tat-Seng},
  booktitle={Findings of the Association for Computational Linguistics: EMNLP 2023},
  year={2023}
}

@inproceedings{hu2024uncertainty,
  title={Uncertainty of Thoughts: Uncertainty-Aware Planning Enhances Information Seeking in Large Language Models},
  author={Hu, Zhiyuan and Liu, Chumin and Feng, Xidong and Zhao, Yilun and Ng, See-Kiong and Luu, Anh Tuan and He, Junxian and Koh, Pang Wei and Hooi, Bryan},
  booktitle={Advances in Neural Information Processing Systems},
  year={2024}
}

@inproceedings{madras2018predict,
  title={Predict Responsibly: Improving Fairness and Accuracy by Learning to Defer},
  author={Madras, David and Pitassi, Toni and Zemel, Richard},
  booktitle={Advances in Neural Information Processing Systems},
  year={2018}
}

@inproceedings{ouyang2022training,
  title={Training Language Models to Follow Instructions with Human Feedback},
  author={Ouyang, Long and Wu, Jeffrey and Jiang, Xu and Almeida, Diogo and Wainwright, Carroll L. and Mishkin, Pamela and Zhang, Chong and Agarwal, Sandhini and Slama, Katarina and Ray, Alex and Schulman, John and Hilton, Jacob and Kelton, Fraser and Miller, Luke and Simens, Maddie and Askell, Amanda and Welinder, Peter and Christiano, Paul and Leike, Jan and Lowe, Ryan},
  booktitle={Advances in Neural Information Processing Systems},
  year={2022}
}

@inproceedings{ma2024spatialpin,
  title={SpatialPIN: Enhancing Spatial Reasoning Capabilities of Vision-Language Models through Prompting and Interacting 3D Priors},
  author={Ma, Chenyang and Lu, Kai and Cheng, Ta-Ying and Trigoni, Niki and Markham, Andrew},
  booktitle={Proceedings of the Conference on Neural Information Processing Systems (NeurIPS)},
  year={2024}
}
\clearpage
\appendix

\section{Related Work}

\textbf{Long-Term Human-Robot Collaboration.}
Prior research has studied collaboration in a range of relatively structured settings, including text-based interaction, dialogue, cooperative games, and other task-specific environments~\citep{vaccaro2024combinations,dhillon2024shaping,ma2024spatialpin,hwang202580,wan2024felt,skantze2025applying,cao2026reframing,gessler2025overcookedv2,liang2024learning,zhu2025multiagentbench}. These settings are valuable for analyzing coordination strategies and interaction patterns, but they typically abstract away key challenges of long-term embodied human-robot collaboration, such as continual perception, physical grounding, and adaptation to a human partner over repeated interactions. Recent advances in embodied simulators and benchmarks have enabled the study of richer collaborative behaviors in realistic 3D environments~\citep{chang2024partnr,yang2025embodiedbench,ramrakhya2025grounding}. However, much of this work still focuses on relatively short-horizon, episodic, or pre-specified tasks, leaving cross-day collaboration across repeated interactions less explored. In contrast, PACT studies embodied collaboration across multiple days, where a robot must progressively accumulate partner-specific knowledge and use this evolving understanding to provide assistance over time.

\textbf{Structured Reasoning for Assistance.}
Recent work has increasingly explored structured reasoning for embodied assistance, where agents use language instructions, scene observations, and contextual cues to guide action selection~\citep{lu2025thinkbot,monwilliams2025ellmer,wang2025affordbot}. More broadly, recent advances in language-model reasoning suggest that structured planning, multi-level reasoning, and adaptive information control can improve reliability in complex decision-making problems~\citep{xiong2025deliberate,xiong2026enhancing,xiong2026scaling}. For example, Situated Instruction Following (SIF) studies ambiguous and temporally evolving instructions grounded in human actions and environmental dynamics, moving beyond literal instruction execution~\citep{min2024situated}. Relatedly, SIFToM shows that spoken instruction following can benefit from pragmatic reasoning about the speaker's goals and joint plans, especially under noisy inputs~\citep{ying2024siftom}. FISER further introduces explicit intention reasoning before action planning~\citep{wan2025infer}. However, these approaches mainly focus on ambiguity resolution within the current interaction or task episode. In contrast, PACT studies structured reasoning in cross-day collaboration, where agents must progressively build and update models of their human partners across repeated interactions in order to infer context-dependent intentions and collaborative task needs over time.

\textbf{Proactive Clarification in Human-Agent Collaboration.}
Recent work has studied proactive clarification as a mechanism for resolving ambiguity and missing information in human-agent collaboration, where agents decide whether to act directly or seek additional information before acting~\citep{taioli2025collaborative,yu2025mixed}. In embodied settings, some approaches enable agents to request expert assistance as part of task execution~\citep{singh2022ask4help}, while others emphasize uncertainty-aware help-seeking for LLM-based planners by aligning model uncertainty with when robots should ask for help. Relatedly, recent dialogue work suggests that effective clarification should be guided by model uncertainty rather than merely imitating human questioning behavior~\citep{testoni2024asking}. However, these studies mainly examine clarification within a single interaction or task episode. In contrast, PACT studies proactive clarification in cross-day human-robot collaboration, where agents must progressively build partner-specific understanding over repeated interactions and use this evolving knowledge to decide when clarification is necessary before acting.

\section{Additional Method Details}
\label{app:additional_method_details}

\subsection{Additional Details of Proactive Ask-or-Act Inference}
\label{app:ask_or_act_details}

This section provides additional details for the proactive ask-or-act procedure introduced in Sec.~\ref{sec:ask_or_act}. In particular, it expands the target-specific heads \(\mathrm{Head}_m\) and inference update \(\mathrm{Infer}_m\) used in Algorithm~\ref{alg:pact}. PACT applies ask-or-act inference to two targets: the human's intent and the concrete task need. The intent target captures what the human currently intends to accomplish, while the task target captures the specific assistance need, target object, or constraint required to support that intent.

\paragraph{Backbone encoding.}
The multimodal large language model \(F_\theta\) introduced in Sec.~\ref{sec:ask_or_act} produces a shared backbone representation used by all heads. Given the robot-observable interaction state \(x_t=(o_t,c_t,H_{<t})\), where \(o_t\) denotes the current multimodal observations, \(c_t\) denotes the temporal context, and \(H_{<t}\) denotes the accumulated cross-day history, \(F_{\theta}\) produces:
\begin{equation}
    z_t = F_{\theta}(o_t,c_t,H_{<t}).
    \label{eq:app_backbone_encoding}
\end{equation}
This representation is used inside the intent head, task head, and ask head. For readability, Algorithm~\ref{alg:pact} abstracts this internal use of \(z_t\) into the target-specific head \(\mathrm{Head}_m\).

\paragraph{Intent inference and clarification.}
For intent inference, the target-specific context is the robot-observable interaction state:
\begin{equation}
    \mathcal{C}_t^{\mathrm{intent}} = x_t .
\end{equation}
The intent head generates a set of candidate human intents:
\begin{equation}
    \mathcal{Y}_t^{\mathrm{intent}}
    =
    \{y_{t,1}^{\mathrm{intent}},\ldots,y_{t,N_I}^{\mathrm{intent}}\}
    =
    f_{\mathrm{intent}}(z_t,\mathcal{C}_t^{\mathrm{intent}}),
    \label{eq:app_intent_candidates}
\end{equation}
where each \(y_{t,i}^{\mathrm{intent}}\) is a candidate description of the human's current intent. The intent head then scores and filters these candidates:
\begin{equation}
    s_{t,i}^{\mathrm{intent}}
    =
    g_{\mathrm{intent}}
    (y_{t,i}^{\mathrm{intent}},z_t,\mathcal{C}_t^{\mathrm{intent}}),
\end{equation}
\begin{equation}
    \widehat{\mathcal{Y}}_t^{\mathrm{intent}}
    =
    \mathrm{Filter}_{\mathrm{intent}}
    \bigl(
    \mathcal{Y}_t^{\mathrm{intent}},
    \{s_{t,i}^{\mathrm{intent}}\}_{i=1}^{N_I}
    \bigr).
    \label{eq:app_intent_filtering}
\end{equation}
Thus, \(\mathrm{Head}_{\mathrm{intent}}(\mathcal{C}_t^{\mathrm{intent}})\) in Algorithm~\ref{alg:pact} corresponds to Eqs.~\eqref{eq:app_intent_candidates}--\eqref{eq:app_intent_filtering}, with internal use of the backbone representation \(z_t\). The set \(\widehat{\mathcal{Y}}_t^{\mathrm{intent}}\) denotes the inferred intents retained after candidate generation, scoring, and filtering.

The ask head estimates whether intent clarification is needed:
\begin{equation}
    q_t^{\mathrm{intent}}
    =
    \pi_{\mathrm{ask}}^{\mathrm{intent}}
    (x_t,
    \widehat{\mathcal{Y}}_t^{\mathrm{intent}},
    b_t),
    \qquad
    q_t^{\mathrm{intent}}\in\{0,1\}.
    \label{eq:app_ask_intent}
\end{equation}
Here \(b_t\) serves as a soft signal to the policy; when \(b_t=0\), the hard budget constraint in Algorithm~\ref{alg:pact} enforces \(q_t^{\mathrm{intent}}=0\). If \(q_t^{\mathrm{intent}}=1\), the robot asks an intent clarification question and receives a human response \(r_t^{\mathrm{intent}}\); otherwise, \(r_t^{\mathrm{intent}}=\varnothing\). The final inferred intents are computed as
\begin{equation}
    \widetilde{\mathcal{Y}}_t^{\mathrm{intent}}
    =
    \mathrm{Infer}_{\mathrm{intent}}
    (
    \widehat{\mathcal{Y}}_t^{\mathrm{intent}},
    \mathcal{C}_t^{\mathrm{intent}},
    r_t^{\mathrm{intent}}
    ).
    \label{eq:app_final_intent}
\end{equation}
When \(r_t^{\mathrm{intent}}=\varnothing\), \(\mathrm{Infer}_{\mathrm{intent}}\) retains \(\widehat{\mathcal{Y}}_t^{\mathrm{intent}}\) as the final result. When \(r_t^{\mathrm{intent}}\neq\varnothing\), the final inferred intents additionally incorporate the human response.

\paragraph{Task inference and clarification.}
Task inference is performed after intent inference because the concrete task need depends on the inferred high-level intent. The target-specific context for task inference is
\begin{equation}
    \mathcal{C}_t^{\mathrm{task}}
    =
    (x_t,\widetilde{\mathcal{Y}}_t^{\mathrm{intent}}),
\end{equation}
where \(\widetilde{\mathcal{Y}}_t^{\mathrm{intent}}\) are the final inferred intents from the previous stage. Conditioned on the final inferred intents, the task head generates candidate task needs:
\begin{equation}
    \mathcal{Y}_t^{\mathrm{task}}
    =
    \{y_{t,1}^{\mathrm{task}},\ldots,y_{t,N_T}^{\mathrm{task}}\}
    =
    f_{\mathrm{task}}(z_t,\mathcal{C}_t^{\mathrm{task}}),
    \label{eq:app_task_candidates}
\end{equation}
where each \(y_{t,j}^{\mathrm{task}}\) denotes a candidate task need, target object, or assistance requirement. The task head then scores and filters these candidates:
\begin{equation}
    s_{t,j}^{\mathrm{task}}
    =
    g_{\mathrm{task}}
    (y_{t,j}^{\mathrm{task}},z_t,\mathcal{C}_t^{\mathrm{task}}),
\end{equation}
\begin{equation}
    \widehat{\mathcal{Y}}_t^{\mathrm{task}}
    =
    \mathrm{Filter}_{\mathrm{task}}
    \bigl(
    \mathcal{Y}_t^{\mathrm{task}},
    \{s_{t,j}^{\mathrm{task}}\}_{j=1}^{N_T}
    \bigr).
    \label{eq:app_task_filtering}
\end{equation}
Thus, \(\mathrm{Head}_{\mathrm{task}}(\mathcal{C}_t^{\mathrm{task}})\) in Algorithm~\ref{alg:pact} corresponds to Eqs.~\eqref{eq:app_task_candidates}--\eqref{eq:app_task_filtering}, with internal use of the backbone representation \(z_t\). The set \(\widehat{\mathcal{Y}}_t^{\mathrm{task}}\) denotes the inferred task needs retained after candidate generation, scoring, and filtering.

The ask head estimates whether task clarification is needed:
\begin{equation}
    q_t^{\mathrm{task}}
    =
    \pi_{\mathrm{ask}}^{\mathrm{task}}
    (x_t,
    \widehat{\mathcal{Y}}_t^{\mathrm{task}},
    b_t),
    \qquad
    q_t^{\mathrm{task}}\in\{0,1\}.
    \label{eq:app_ask_task}
\end{equation}
As with intent clarification, \(b_t\) serves as a soft signal, while the hard budget constraint in Algorithm~\ref{alg:pact} enforces \(q_t^{\mathrm{task}}=0\) when \(b_t=0\). If \(q_t^{\mathrm{task}}=1\), the robot asks a task clarification question and receives a human response \(r_t^{\mathrm{task}}\); otherwise, \(r_t^{\mathrm{task}}=\varnothing\). The final inferred task needs are
\begin{equation}
    \widetilde{\mathcal{Y}}_t^{\mathrm{task}}
    =
    \mathrm{Infer}_{\mathrm{task}}
    (
    \widehat{\mathcal{Y}}_t^{\mathrm{task}},
    \mathcal{C}_t^{\mathrm{task}},
    r_t^{\mathrm{task}}
    ).
    \label{eq:app_final_task}
\end{equation}
When \(r_t^{\mathrm{task}}=\varnothing\), \(\mathrm{Infer}_{\mathrm{task}}\) retains \(\widehat{\mathcal{Y}}_t^{\mathrm{task}}\) as the final result. When \(r_t^{\mathrm{task}}\neq\varnothing\), the final inferred task needs additionally incorporate the human response.

\paragraph{Action execution and history update.}
After obtaining the final inferred intents and task needs, the robot executes an assistance action:
\begin{equation}
    a_t =
    \pi_{\mathrm{act}}
    (
    x_t,
    \widetilde{\mathcal{Y}}_t^{\mathrm{intent}},
    \widetilde{\mathcal{Y}}_t^{\mathrm{task}}
    ).
    \label{eq:app_action}
\end{equation}
After execution, the robot observes the action outcome and feedback \(\phi_t\), and appends the interaction record to the cross-day history:
\begin{equation}
    H_t =
    H_{<t}
    \cup
    \{
    x_t,
    \widetilde{\mathcal{Y}}_t^{\mathrm{intent}},
    \widetilde{\mathcal{Y}}_t^{\mathrm{task}},
    q_t^{\mathrm{intent}},
    q_t^{\mathrm{task}},
    r_t^{\mathrm{intent}},
    r_t^{\mathrm{task}},
    a_t,
    \phi_t
    \}.
    \label{eq:app_history_update}
\end{equation}
This updated history is retrieved in subsequent interaction steps and days.

\paragraph{Clarification responses and offline labels.}
Clarification responses are only available after the robot explicitly asks. In our simulator, the simulated human has a latent state \(\xi_t\), which includes the underlying intent and task need. This latent state is not exposed to the robot as online input. Instead, it is used by the simulated human to generate a target-specific response \(r_t^m\) when \(q_t^m=1\), where \(m\in\{\mathrm{intent},\mathrm{task}\}\). If the robot does not ask, no clarification response is returned.

We distinguish online clarification responses from offline labels and end-of-day feedback. Online clarification responses are part of the interaction: they are observed only after the corresponding ask-or-act output is \(1\), consume the ask budget, and can affect the robot's inference within the current step. Offline labels and end-of-day feedback are used to assess whether the robot's inferred intent and task assistance matched the human's actual needs. They are used to derive training and evaluation signals for subsequent days, but are not available to the robot during online inference.

\begin{table}[t]
\centering
\caption{Clarification targets in PACT.}
\label{tab:ask_or_act_targets_app}
\small
\begin{tabular}{p{0.14\linewidth} p{0.32\linewidth} p{0.22\linewidth} p{0.22\linewidth}}
\toprule
Target \(m\) & Context & Candidate set & Clarification response \\
\midrule
Intent
& \(\mathcal{C}_t^{\mathrm{intent}}=x_t\)
& Candidate human intents
& Clarifies the human's current intent \\
Task
& \(\mathcal{C}_t^{\mathrm{task}}=(x_t,\widetilde{\mathcal{Y}}_t^{\mathrm{intent}})\)
& Candidate task needs
& Clarifies the required task need, target object, or constraint \\
\bottomrule
\end{tabular}
\end{table}

\subsection{Additional Details of Clarification Utility}
\label{app:clarification_utility}

This section provides additional details for the Clarification Utility metric introduced in Section~\ref{sec:clarification_utility}. Let \(T\) be the total number of interaction steps, \(A_t\in[0,1]\) be the post-action assistance score at step \(t\), and \(q_t\) be the number of clarification questions asked at step \(t\). The total number of clarification questions is
\begin{equation}
    K = \sum_{t=1}^{T} q_t .
\end{equation}
Clarification Utility is defined as
\begin{equation}
    U = \frac{\sum_{t=1}^{T} A_t}{T+K}.
\end{equation}

To connect this definition with standard evaluation quantities, define the average assistance score and ASK rate as
\begin{equation}
    \mathrm{Acc} = \frac{1}{T}\sum_{t=1}^{T} A_t,
    \qquad
    \mathrm{AskRate} = \frac{K}{T}.
\end{equation}
Here, \(\mathrm{AskRate}\) measures the average number of clarification questions per interaction step. Then Clarification Utility can be rewritten as
\begin{equation}
    U
    = \frac{\sum_{t=1}^{T} A_t}{T+K}
    = \frac{\frac{1}{T}\sum_{t=1}^{T} A_t}{1+\frac{K}{T}}
    = \frac{\mathrm{Acc}}{1+\mathrm{AskRate}}.
    \label{eq:clarification_utility_rewrite}
\end{equation}

This form shows two basic properties. First, when two methods achieve the same average assistance score, the method with fewer clarification questions obtains higher utility. Specifically, for two methods with the same average score \(a\) and ASK rates \(r_A<r_B\),
\begin{equation}
    U_A=\frac{a}{1+r_A}
    >
    \frac{a}{1+r_B}
    =U_B .
\end{equation}

Second, asking more questions is beneficial only when it leads to a sufficient improvement in assistance performance. For two methods with average assistance scores \(a_A,a_B\) and ASK rates \(r_A,r_B\), method \(B\) has higher utility than method \(A\) only if
\begin{equation}
    \frac{a_B}{1+r_B}
    >
    \frac{a_A}{1+r_A}.
\end{equation}
Equivalently,
\begin{equation}
    a_B
    >
    a_A \frac{1+r_B}{1+r_A}.
\end{equation}
Thus, a higher ASK rate must be justified by a sufficiently large gain in assistance performance.

In the special case where step-level assistance score is binary, \(A_t\in\{0,1\}\), let
\begin{equation}
    C = \sum_{t=1}^{T} A_t
\end{equation}
be the number of successful assistance steps. The metric then reduces to the count-based form
\begin{equation}
    U = \frac{C}{T+K}.
\end{equation}
Thus, the count-based definition is a special case of the more general score-based Clarification Utility.

Finally, since \(A_t\in[0,1]\) and \(K\geq 0\), Clarification Utility is bounded by
\begin{equation}
    0 \leq U \leq 1.
\end{equation}
When \(K=0\), the metric reduces to the average assistance score, \(U=\mathrm{Acc}\). When \(K>0\), the metric penalizes additional clarification unless it improves the final assistance outcome.

\section{Additional Experimental Details}
\label{app:additional_experimental_details}

\subsection{Environment and Simulated-Human Details}
\label{app:environment_simulated_human_details}

\noindent\textbf{Simulated-human construction.}
PACT evaluates proactive asking with profile-conditioned simulated human partners. 
The simulated-human construction is based on the continual HRC protocol introduced in COOPERA~\citep{ma2025coopera}, where simulated humans are driven by persistent psychological traits, long-term routines, temporally evolving intentions, and accumulated interaction history. 
At each hourly interaction step, the simulated human generates behavior conditioned on its profile, current time, scene context, and retrieved past behaviors. 
This design allows the robot to observe evolving human behavior over multiple days and infer assistance needs without receiving explicit task commands at every step.

\noindent\textbf{Simulation sources and generation pipeline.}
Table~\ref{tab:human_simulation_pipeline} summarizes the main components used to instantiate simulated humans in our experiments. 
The simulated humans are built from synthetic profiles with psychometric attributes, placed in dynamic household environments, and grounded with embodied motion data. 
The behavior generator and memory retriever are used to produce temporally dependent human intentions and tasks across days.

\begin{table}[t]
\centering
\small
\caption{\textbf{Simulated-human construction pipeline.}}
\label{tab:human_simulation_pipeline}
\renewcommand{\arraystretch}{1.12}
\begin{tabularx}{\linewidth}{@{}>{\raggedright\arraybackslash}p{0.22\linewidth}
>{\raggedright\arraybackslash}p{0.27\linewidth}
>{\raggedright\arraybackslash}X@{}}
\toprule
\rowcolor{PACTHeader}
\textbf{Component} & \textbf{Implementation} & \textbf{Role in simulation} \\
\midrule
Simulation platform & Habitat 3.0~\citep{habitat3} & Embodied household simulation. \\
\rowcolor{PACTRow}
Household scenes & HSSD~\citep{hssd200} & 3D indoor environments with household objects. \\
Dynamic objects & YCB objects~\citep{ycb} and movable HSSD objects & Object interaction and persistent state changes across days. \\
\rowcolor{PACTRow}
Robot embodiment & Fetch mobile manipulator~\citep{fetch_freight} & Robot assistant used in the household environment. \\
Human profiles & SPC synthetic profiles & Persistent user descriptions and psychometric attributes. \\
\rowcolor{PACTRow}
Personality attributes & Big-Five traits & Trait-conditioned behavioral variation across simulated humans. \\
Human behavior model & Llama-3.1-8B, temperature 0.7 & Hourly intention and task generation. \\
\rowcolor{PACTRow}
Memory retrieval & MiniLM-L6-v2~\citep{wang2020minilm}, \(\lambda=0.95\) & Retrieval of relevant past interaction history. \\
Retrieved memory & Relevant past intentions and tasks & Context for temporally dependent behavior generation. \\
\rowcolor{PACTRow}
Motion sources & Motion-X~\citep{lin2023motionx} and AMASS~\citep{mahmood2019amass} & Embodied whole-body human motion. \\
Number of profiles & 10 simulated humans & Diverse human partners across collaboration settings. \\
\bottomrule
\end{tabularx}
\end{table}

\noindent\textbf{Environment scale.}
The simulated humans interact in dynamic household scenes. 
The environment contains static household objects from HSSD and additional movable objects from selected object categories and YCB. 
Across the selected scenes, the benchmark covers hundreds of rooms and thousands of static and dynamic objects, allowing human behavior and robot assistance to depend on both persistent human traits and changing scene states.

\begin{table}[t]
\centering
\small
\caption{\textbf{Household environment statistics used for simulated-human interaction.}}
\label{tab:human_simulation_environment_stats}
\renewcommand{\arraystretch}{1.12}
\begin{tabular*}{0.82\linewidth}{@{\extracolsep{\fill}}lr@{}}
\toprule
\rowcolor{PACTHeader}
\textbf{Environment statistic} & \textbf{Value} \\
\midrule
Static objects in HSSD & 18,656 \\
\rowcolor{PACTRow}
Selected household scenes & 5 \\
Total rooms in selected scenes & 411 \\
\rowcolor{PACTRow}
Total static objects in selected scenes & 51,140 \\
Total dynamic objects in selected scenes & 3,394 \\
\rowcolor{PACTRow}
Additional YCB dynamic objects & 20 \\
\bottomrule
\end{tabular*}
\end{table}

\noindent\textbf{Validation with real-user routines.}
PACT uses an existing profile-conditioned simulated-human protocol whose behavioral validity has been previously evaluated. 
Tables~\ref{tab:human_simulation_validation} and~\ref{tab:human_simulation_real_alignment} summarize the reported validation evidence. 
The validation covers whether simulated humans are distinguishable by generated intentions and tasks, whether their personality traits are diverse, whether generated behaviors align with psychometric attributes, whether hourly intentions exhibit temporal dependence, whether real users can identify consistent simulated humans, and whether profile-conditioned simulated intentions align with real-human daily intentions.

\begin{table}[t]
\centering
\small
\caption{\textbf{Validation results for the simulated-human protocol.}}
\label{tab:human_simulation_validation}
\renewcommand{\arraystretch}{1.12}
\begin{tabularx}{\linewidth}{@{}>{\raggedright\arraybackslash}p{0.20\linewidth}
>{\raggedright\arraybackslash}X
>{\raggedright\arraybackslash}p{0.24\linewidth}
r@{}}
\toprule
\rowcolor{PACTHeader}
\textbf{Validation aspect} & \textbf{Evaluation goal} & \textbf{Metric} & \textbf{Value} \\
\midrule
Human classification & Different simulated humans produce distinguishable intention patterns. & Intention classification accuracy & 0.995 \\
\rowcolor{PACTRow}
Human classification & Different simulated humans produce distinguishable task preferences. & Task classification accuracy & 0.830 \\
Simulated-human diversity & Generated humans cover diverse personality profiles. & Big-Five trait standard deviation & 0.939 \\
\rowcolor{PACTRow}
Trait--psychometric coherence & Behavior is positively correlated with the matched psychometric trait. & Aligned correlation & 0.342 \\
Trait--psychometric coherence & Behavior is negatively correlated with mismatched traits. & Mismatched correlation & -0.497 \\
\rowcolor{PACTRow}
Temporal dependence & Current intentions depend on previous within-day intentions. & Prediction accuracy & 0.789 \\
Temporal dependence & Temporal prediction remains balanced under F1. & F1 & 0.790 \\
\rowcolor{PACTRow}
User studies & Real users can identify the same simulated human across behaviors. & Same-human identification accuracy & 0.764 \\
User studies & Real users can match behaviors to trait descriptions. & Trait--behavior matching accuracy & 0.712 \\
\bottomrule
\end{tabularx}
\end{table}

\noindent\textbf{Interpretation of validation results.}
The human-classification results indicate that different simulated humans produce distinguishable intention and task patterns. 
The diversity score shows that the generated humans cover a broad range of Big-Five traits. 
The coherence results compare aligned and mismatched human--trait pairs: the aligned correlation is positive, while the mismatched correlation is negative, supporting the relationship between generated behavior and psychometric attributes. 
The temporal-dependence results indicate that current intentions are predictable from earlier intentions within the day. 
The user-study results further show that real participants can identify the same simulated human across generated behaviors and match behavioral traces to trait descriptions above chance.

\noindent\textbf{Alignment with real-human routines.}
In addition to the validation above, the protocol compares simulated intentions with real-human intentions collected from participants over multiple days. 
Table~\ref{tab:human_simulation_real_alignment} summarizes the reported semantic alignment results. 
The profile-conditioned simulation achieves higher semantic similarity than generic prompting or mismatched human--simulation pairs, suggesting that conditioning on human traits improves alignment with real-human daily routines.

\begin{table}[t]
\centering
\small
\caption{\textbf{Semantic alignment between simulated and real-human intentions.}}
\label{tab:human_simulation_real_alignment}
\renewcommand{\arraystretch}{1.12}
\begin{tabular}{@{}lccc@{}}
\toprule
\rowcolor{PACTHeader}
\textbf{Embedding metric} & \textbf{Generic} & \textbf{Mismatched} & \textbf{Profile-conditioned} \\
\midrule
SBERT & 0.554 & 0.523 & 0.810 \\
\rowcolor{PACTRow}
OpenAI Embedding & 0.537 & 0.543 & 0.772 \\
\bottomrule
\end{tabular}
\end{table}

These validation results support the use of profile-conditioned simulated humans as interaction partners in PACT. 
Therefore, our experiments focus on proactive clarification and ask-or-act decision making under this validated simulated-human protocol.

\subsection{Collaboration Types and Settings}
\label{app:collaboration_settings}

\noindent\textbf{Collaboration types.}
We evaluate PACT on two collaboration types with increasing task complexity and openness. 
Table~\ref{tab:collaboration_types} summarizes their main differences.

\begin{table}[t]
\centering
\small
\caption{\textbf{Collaboration types used in PACT.}}
\label{tab:collaboration_types}
\renewcommand{\arraystretch}{1.12}
\begin{tabularx}{\linewidth}{@{}p{0.16\linewidth}p{0.28\linewidth}p{0.25\linewidth}X@{}}
\toprule
\rowcolor{PACTHeader}
\textbf{Type} & \textbf{Task structure} & \textbf{Robot observation} & \textbf{Robot assistance target} \\
\midrule
Type 1 & 
Each intent is decomposed into three pick-and-place tasks. & 
First-task video with a textual description. & 
Infer and assist with the remaining tasks based on objects available in the scene. \\
\rowcolor{PACTRow}
Type 2 & 
Each intent is decomposed into five free-form human-motion tasks involving static objects. & 
First-task video without textual guidance. & 
Infer useful assistance beyond fixed pick-and-place actions; the robot may propose any helpful object or support action. \\
\bottomrule
\end{tabularx}
\end{table}

\noindent\textbf{Evaluation settings.}
For both collaboration types, we evaluate PACT under four day-spanning human--scene settings. 
These settings vary whether the human partner and scene remain fixed or change across days, as summarized in Table~\ref{tab:collaboration_setting_details}.

\begin{table}[t]
\centering
\small
\caption{\textbf{Day-spanning human--scene evaluation settings.}}
\label{tab:collaboration_setting_details}
\renewcommand{\arraystretch}{1.12}
\begin{tabularx}{\linewidth}{@{}p{0.24\linewidth}p{0.17\linewidth}p{0.17\linewidth}X@{}}
\toprule
\rowcolor{PACTHeader}
\textbf{Setting} & \textbf{Rollout} & \textbf{Scene} & \textbf{Evaluation focus} \\
\midrule
Same human / same scene & 
5 days & 
1 scene & 
Repeated interaction with one consistent partner in a fixed environment. \\
\rowcolor{PACTRow}
Same human / different scenes & 
5 days & 
5 scenes & 
Transfer of partner-specific history across daily scene changes. \\
Different humans / same scene & 
9 days & 
1 scene & 
Rotation over Human 1, Human 2, and Human 3 for three rounds in the same scene. \\
\rowcolor{PACTRow}
Different humans / different scenes & 
9 days & 
3 scenes & 
Rotation over Human 1, Human 2, and Human 3 across three scenes, testing joint human--scene variation. \\
\bottomrule
\end{tabularx}
\end{table}

Together, these settings evaluate repeated interaction with a consistent partner, adaptation across environments, transfer across human partners, and generalization under joint human--scene variation. 
The two cross-human settings also examine whether the robot can benefit from transferable interaction patterns when the history available for each individual human is limited.

\noindent\textbf{Rollout structure.}
All methods follow the same multi-day rollout protocol described in Sec.~\ref{sec:exp_setup}. 
The same-human settings contain five-day sequences, while the cross-human settings contain nine day--human segments. 
For downstream collaboration performance, we report task-level F1 and intent-level F1 in the main paper and provide additional accuracy-based analyses in Appendix~\ref{app:additional_global_metrics}. 
The clarification-specific metrics are defined in Sec.~\ref{sec:clarification_utility}.

\subsection{Model, Training, and Compute Details}
\label{app:model_training_details}

\noindent\textbf{Robot reasoning backbone.}
For robot-side visual-language reasoning, we use Qwen3.5-27B as the shared multimodal reasoning backbone and serve it with vLLM through an OpenAI-compatible inference interface.
At each interaction step, the backbone receives the robot-observable context, including current multi-view observations, temporal context, scene information, candidate inferences, remaining ask budget, and accumulated interaction history.
This shared context is used for intent inference, task inference, and ask-or-act decision making.

\noindent\textbf{PACT head instantiation.}
For learnable variants, the intent head, task head, and ask head in Fig.~\ref{fig:framework} are implemented with Qwen3-8B-based classifier heads.
The intent head filters candidate human intentions, and the task head filters candidate task predicates or assistance needs.
The ask head predicts whether clarification is needed at the intent or task level.
At the intent level, it estimates whether the robot should ask about the human's high-level intention.
At the task level, it estimates whether the robot should ask about the concrete task need, target object, or constraint.
Prompting-based variants use the same robot-observable context but determine whether to ask through prompting, without updating ask-specific classifier heads.

\noindent\textbf{Online inference procedure.}
For each day \(d\) and interaction step \(t\), the simulated human first produces the latent ground-truth intention and task predicates according to the profile-conditioned human protocol.
These latent states are not exposed to the robot during online inference unless the robot asks for clarification.
The robot then performs intent inference, task inference, and optional clarification before action execution.

First, the robot discovers candidate intentions from the current observation and accumulated history.
The intent head filters these candidates into selected intentions.
The ask head then predicts whether intent-level clarification is needed.
If the output is \textsc{NoAsk}, the robot keeps the selected intention list.
If the output is \textsc{Ask}, the robot asks the simulated human for clarification and replaces the execution-time intention with the clarified ground-truth intention.

Conditioned on the final execution-time intention, the robot then discovers candidate task predicates.
The task head filters these predicates into selected task needs.
The ask head predicts whether task-level clarification is needed.
If the output is \textsc{NoAsk}, the robot keeps the selected task predicates.
If the output is \textsc{Ask}, the robot receives the required task, object, or constraint information from the simulated human and uses it to correct the current assistance output.

\noindent\textbf{Offline supervision and day-end update.}
After each interaction step, an offline human-judge evaluation produces supervision signals for training and analysis.
The judge evaluates whether the selected intentions, task predicates, and object or category choices match the simulated human's latent needs.
These signals are used to derive intent approval, predicate approval, category approval, and intent-level and task-level ask labels.
During each day, we collect training data for intention filtering, predicate filtering, and ask-head training.
At the end of the day, training-based variants update the corresponding LoRA modules using the accumulated data from that day.
The updated heads are then used on the next day, allowing the robot to adapt from previous interaction history and feedback.

\noindent\textbf{Training and compute configuration.}
Table~\ref{tab:model_training_config} summarizes the model, training, and compute configuration used in our experiments.

\begin{table}[t]
\centering
\small
\caption{\textbf{Model, training, and compute configuration.}}
\label{tab:model_training_config}
\renewcommand{\arraystretch}{1.12}
\begin{tabularx}{0.94\linewidth}{@{}p{0.34\linewidth}X@{}}
\toprule
\rowcolor{PACTHeader}
\textbf{Item} & \textbf{Configuration} \\
\midrule
Robot reasoning backbone & Qwen3.5-27B served with vLLM through an OpenAI-compatible interface \\
\rowcolor{PACTRow}
Learnable head backbone & Qwen3-8B \\
Learnable heads & Intent head, task head, and ask head \\
\rowcolor{PACTRow}
Intent/task head role & Candidate filtering for intentions and task predicates \\
Ask head role & Binary clarification prediction at the intent and task levels \\
\rowcolor{PACTRow}
Prompting-based variants & Use the same robot-observable context but do not update ask-specific classifier heads \\
Training-based variants & Updated at the end of each day using accumulated interaction data and offline supervision signals \\
\rowcolor{PACTRow}
Finetuning method & LoRA applied to query, key, value, and output projection modules \\
LoRA configuration & Rank 8, alpha 16, dropout 0.2 \\
\rowcolor{PACTRow}
Optimizer & AdamW \\
Learning rate & \(1\times10^{-5}\) \\
\rowcolor{PACTRow}
Weight decay & 0.01 \\
Training epochs & 5 \\
\rowcolor{PACTRow}
Batch size & 2 \\
Gradient accumulation & 4 steps \\
\rowcolor{PACTRow}
Compute resources & Three NVIDIA RTX A6000 GPUs, each with 48GB memory \\
\bottomrule
\end{tabularx}
\end{table}

\definecolor{PACTTableHeader}{RGB}{226,232,242}
\definecolor{PACTTableRow}{RGB}{248,250,253}

\section{Implementation Details of ASK Strategies}
\label{app:ask_strategies}

All ASK strategies are inserted into the same PACT decision pipeline.
Before an ASK decision is made, the robot first generates multiple candidate intents and filters them with the learned intent classifier.
For each retained intent, it then generates task or predicate candidates, including the associated object or action information when available, and filters them with the learned task classifier.
The ASK module does not change this pre-ASK candidate generation and filtering process.
It only decides whether the robot should ask the human for clarification at two points: the intent stage and the task stage.
If an intent-stage query is triggered, the robot uses the human-confirmed intent for the downstream task planning step.
If a task-stage query is triggered, the robot uses the human-confirmed tasks and object/action choices before execution.

Among the learned ASK strategies, ASK-RL is our primary PACT instantiation.
ASK-SFT and ASK-L2D use the same ASK decision interface but differ in their training objective.
The prompt-only strategies use the same pre-ASK state but do not train or load ASK LoRA adapters and do not update an RL policy.
Table~\ref{tab:ask_strategy_summary} summarizes the eight ASK strategies used in our comparison.

\begin{table}[t]
\centering
\caption{ASK strategy configuration.}
\label{tab:ask_strategy_summary}
\scriptsize
\setlength{\tabcolsep}{2pt}
\renewcommand{\arraystretch}{1.08}
\rowcolors{2}{PACTTableRow}{white}
\begin{tabularx}{\linewidth}{
p{0.20\linewidth}
p{0.16\linewidth}
>{\raggedright\arraybackslash}X
>{\raggedright\arraybackslash}X
>{\raggedright\arraybackslash}X
}
\toprule
\rowcolor{PACTTableHeader}
\textbf{Method} & \textbf{Decision form} & \textbf{Learning signal} & \textbf{Update schedule} & \textbf{Inference style} \\
\midrule
ASK-RL
& Ask-or-act action
& Online reward with ask cost
& End-of-day PPO update
& Policy optimization \\
ASK-SFT
& \textsc{Ask}/\textsc{NoAsk}
& Supervised ASK labels
& End-of-day LoRA update
& Learned classifier \\
ASK-L2D
& Act/defer
& Defer indicator with ask/error costs
& End-of-day LoRA update
& Cost-sensitive learned classifier \\
ASK-ZeroShot
& \textsc{Ask}/\textsc{NoAsk}
& None
& None
& Single-step prompt \\
ASK-FewShot
& \textsc{Ask}/\textsc{NoAsk}
& None
& None
& Single-step prompt with examples \\
ASK-ToT
& \textsc{Ask}/\textsc{NoAsk}
& None
& None
& Tree search \\
ASK-ProactiveCoT
& \textsc{Ask}/\textsc{NoAsk}
& None
& None
& Adaptive sub-question reasoning \\
ASK-UoT
& \textsc{Ask}/\textsc{NoAsk}
& None
& None
& Uncertainty-reduction reasoning \\
\bottomrule
\end{tabularx}
\rowcolors{2}{}{}
\end{table}

\subsection{ASK-RL}

ASK-RL formulates the ASK module as an online control policy.
At each ASK point, the action is
\[
u_t \in \{\textsc{Ask}, \textsc{NoAsk}\}.
\]
A LoRA-backed actor-critic policy scores the two action tokens from the stage-specific ASK prompt.
The intent and task stages maintain independent policies, because the uncertainty sources differ: intent ASK resolves the user's high-level goal, whereas task ASK resolves the concrete task/predicate and object/action choices.

The step-wise reward is:
\[
R_t =
\begin{cases}
1, & \text{if } u_t=\textsc{NoAsk} \text{ and skipping the query is correct},\\
1-c_{\mathrm{ask}}, & \text{if } u_t=\textsc{Ask} \text{ and the query is needed},\\
-1, & \text{if } u_t=\textsc{NoAsk} \text{ and skipping the query is wrong},\\
-c_{\mathrm{ask}}, & \text{if } u_t=\textsc{Ask} \text{ and the query is unnecessary}.
\end{cases}
\]
This reward makes correct autonomous action the best outcome, rewards necessary clarification after subtracting the query cost, strongly penalizes missed queries, and mildly penalizes unnecessary interruptions.
The policy is updated over day-level trajectories using clipped PPO with generalized advantage estimation, value regression, and entropy regularization.
As with ASK-SFT and ASK-L2D, the daily ASK budget is enforced at inference time.

\subsection{ASK-SFT}

ASK-SFT treats the ask-or-act decision as supervised conditional generation.
For each stage, a LoRA-based ASK classifier receives a prompt \(x_t\) describing the current pre-ASK state: candidate intents or tasks, classifier predictions, human profile and traits, interaction history, current time, and the remaining ASK budget.
It predicts
\[
y_t \in \{\textsc{Ask}, \textsc{NoAsk}\}.
\]
The supervision target is derived from judge or outcome feedback.
When the current robot-selected candidates are sufficient for acting without clarification, the target is \textsc{NoAsk}; otherwise, the target is \textsc{Ask}.
The model is trained as a causal language model, but the loss is applied only to the answer tokens:
\[
\mathcal{L}_{\mathrm{SFT}}
=
- \sum_{j \in \mathcal{A}(x_t)}
\log P_{\theta}(y_{t,j} \mid x_t, y_{t,<j}),
\]
where \(\mathcal{A}(x_t)\) denotes the answer-token positions.
Separate ASK classifiers are maintained for the intent and task stages.
At inference time, the classifier outputs \textsc{Ask} or \textsc{NoAsk}; the daily ASK budget is enforced before executing the query.

\subsection{ASK-L2D}

ASK-L2D uses the same stage-wise ASK inputs and LoRA classifier architecture as ASK-SFT, but replaces pure imitation with a learning-to-defer objective.
Let \(e_t \in \{0,1\}\) indicate whether not asking would lead to an incorrect downstream decision.
At the intent stage, this means that the retained intents do not contain a judge-approved intent.
At the task stage, this means that the selected task/predicate set is insufficient, considering both predicate correctness and the associated object/action alignment.

The defer-aware surrogate loss is:
\[
\mathcal{L}_{\mathrm{L2D}}(x_t)
=
c_{\mathrm{err}}\, e_t\, P(\textsc{NoAsk}\mid x_t)
+
c_{\mathrm{ask}}\, P(\textsc{Ask}\mid x_t),
\]
where \(c_{\mathrm{ask}}\) is the cost of interrupting the human and \(c_{\mathrm{err}}\) is the cost of acting incorrectly without asking.
This objective penalizes the model for choosing \textsc{NoAsk} when deferring to the human would have prevented an error, while also charging a fixed cost for unnecessary queries.
At inference, ASK-L2D queries when the estimated probability of asking exceeds the cost-sensitive threshold:
\[
P(\textsc{Ask}\mid x_t) > \frac{c_{\mathrm{ask}}}{c_{\mathrm{err}}}.
\]
Thus, ASK-L2D is explicitly controlled by the trade-off between clarification cost and autonomous-action risk.

\subsection{Single-Step Prompt Strategies}

ASK-ZeroShot and ASK-FewShot are prompt-only ASK baselines.
They use the same pre-ASK state as the learned methods but do not train ASK-specific parameters, do not load ASK LoRA adapters, and do not update an RL policy.
At each decision point, the base language model is queried once and its response is parsed into \(\{\textsc{Ask}, \textsc{NoAsk}\}\).

Let \(x_t\) denote the stage-specific prompt and let \(\mathcal{D}\) denote optional demonstrations.
The two single-step prompt methods are:
\[
\hat{y} =
\begin{cases}
\operatorname{parse}\!\bigl(f_\theta(x_t)\bigr),
& \text{ASK-ZeroShot}, \\
\operatorname{parse}\!\bigl(f_\theta(x_t \oplus \mathcal{D})\bigr),
& \text{ASK-FewShot},
\end{cases}
\]
where \(f_\theta\) is the base language model and \(\operatorname{parse}(\cdot)\) maps the generated text to an ASK decision.

\begin{table}[t]
\centering
\caption{Single-step prompt ASK strategies.}
\label{tab:single_step_prompt_ask}
\small
\setlength{\tabcolsep}{4pt}
\renewcommand{\arraystretch}{1.08}
\rowcolors{2}{PACTTableRow}{white}
\begin{tabularx}{\linewidth}{
p{0.28\linewidth}
p{0.22\linewidth}
>{\raggedright\arraybackslash}X
}
\toprule
\rowcolor{PACTTableHeader}
\textbf{Method} & \textbf{Demonstrations} & \textbf{Decision rule information} \\
\midrule
ASK-ZeroShot & none & Current candidates, classifier results, profile/history, time, and budget \\
ASK-FewShot & two examples & Same rule prompt plus one \textsc{NoAsk} and one \textsc{Ask} example \\
\bottomrule
\end{tabularx}
\rowcolors{2}{}{}
\end{table}

\begin{figure}[t]
\centering
\begin{minipage}{0.96\linewidth}
\small
\begin{tcolorbox}[
  colback=gray!6, colframe=gray!40,
  title={\textbf{Intent-stage ASK prompt} (single-step prompt methods)},
  fonttitle=\small, left=4pt, right=4pt, top=4pt, bottom=4pt
]
\textbf{Task:} Determine whether the robot should ask the human to clarify the true intent.

\medskip
\textbf{Inputs:}
\begin{enumerate}[leftmargin=1.4em, itemsep=0pt, topsep=2pt]
  \item Current time: [\textsc{time}]
  \item Human profile and Big Five traits: [\textsc{profile}, \textsc{traits}]
  \item Relevant previous intents or tasks: [\textsc{history}]
  \item Candidate intents inferred by the robot: [\textsc{intents}]
  \item Classifier predictions for the candidates: [\textsc{yes/no predictions}]
  \item Current ASK count and daily ASK budget: [\textsc{budget}]
\end{enumerate}

\medskip
\textbf{Rule:} Output \textit{``I do not need to ask a question.''} when the predictions are consistent and sufficient for acting; output \textit{``What is your true intent?''} when the predictions are contradictory, inconsistent with the human context, or leave no reliable intent to execute.

\medskip
\textit{ASK-FewShot appends labeled examples before the final query.}
\end{tcolorbox}
\end{minipage}
\caption{Simplified intent-stage prompt structure for the single-step prompt ASK baselines. The task-stage prompt uses the same structure but replaces intents with selected intents, proposed tasks/predicates, object/action information, and predicate classifier results.}
\label{fig:single_step_prompt}
\end{figure}

\subsection{Multi-Step Prompt Strategies}

ASK-ToT, ASK-ProactiveCoT, and ASK-UoT are also prompt-only methods, but they perform multiple model calls before producing the final ASK decision.
They use the same base model wrapper as the single-step prompt methods and do not train ASK-specific parameters.
Table~\ref{tab:multi_step_ask_config} summarizes their reasoning structure.

\begin{table}[t]
\centering
\caption{Configuration of multi-step prompt ASK strategies. Calls are counted per ASK decision under the default setting.}
\label{tab:multi_step_ask_config}
\scriptsize
\setlength{\tabcolsep}{2pt}
\renewcommand{\arraystretch}{1.08}
\rowcolors{2}{PACTTableRow}{white}
\begin{tabularx}{\linewidth}{
p{0.22\linewidth}
p{0.23\linewidth}
>{\raggedright\arraybackslash}X
p{0.18\linewidth}
}
\toprule
\rowcolor{PACTTableHeader}
\textbf{Method} & \textbf{Main setting} & \textbf{Reasoning structure} & \textbf{Typical calls} \\
\midrule
ASK-ProactiveCoT
& \(T=2\) turns
& Select and solve adaptive sub-questions, then summarize
& \(2T+1=5\) \\
ASK-ToT
& \(D=2\), \(B=2\)
& Expand, score, and select reasoning branches
& \(DB{\times}3+1=13\) \\
ASK-UoT
& \(T=2\), 3 candidate checks per turn
& Simulate uncertainty reduction, execute the best check, and update the candidate decision set
& up to \(T(1{+}3{\times}2+2)+1=19\) \\
\bottomrule
\end{tabularx}
\rowcolors{2}{}{}
\end{table}

\begin{algorithm}[t]
\caption{ASK-ProactiveCoT}
\label{alg:ask_proactive_cot}
\begin{algorithmic}[1]
\State Initialize a reasoning record \(\mathcal{R} \leftarrow \emptyset\).
\For{\(t=1\) to \(T\)}
    \State Select a sub-question about the current uncertainty.
    \State Analyze the sub-question using the current pre-ASK state.
    \State Append the result to \(\mathcal{R}\).
\EndFor
\State Summarize \(\mathcal{R}\) and return \textsc{Ask} or \textsc{NoAsk}.
\end{algorithmic}
\end{algorithm}

\begin{algorithm}[t]
\caption{ASK-ToT}
\label{alg:ask_tot}
\begin{algorithmic}[1]
\State Initialize a root reasoning node.
\For{\(d=1\) to \(D\)}
    \For{\(b=1\) to \(B\)}
        \State Generate one candidate analysis branch.
        \State Analyze the ASK decision under that branch.
        \State Score the branch with \(s_{d,b} \in \{0,1,2\}\).
    \EndFor
    \State Keep a highest-scoring branch for the next depth.
\EndFor
\State Summarize the best reasoning path and return \textsc{Ask} or \textsc{NoAsk}.
\end{algorithmic}
\end{algorithm}

\begin{algorithm}[t]
\caption{ASK-UoT}
\label{alg:ask_uot}
\begin{algorithmic}[1]
\State Initialize the candidate decision set \(\mathcal{C}_0 \leftarrow \{\textsc{Ask}, \textsc{NoAsk}\}\).
\For{\(t=1\) to \(T\)}
    \State Generate candidate verification steps for the current uncertainty.
    \State Simulate each step and estimate how much it reduces \(\mathcal{C}_{t-1}\).
    \State Execute the step with the largest estimated reduction.
    \State Update the candidate decision set \(\mathcal{C}_t\).
    \If{\(|\mathcal{C}_t| = 1\)}
        \State \Return the remaining decision.
    \EndIf
\EndFor
\State Resolve the remaining uncertainty and return \textsc{Ask} or \textsc{NoAsk}.
\end{algorithmic}
\end{algorithm}

\begin{figure}[t]
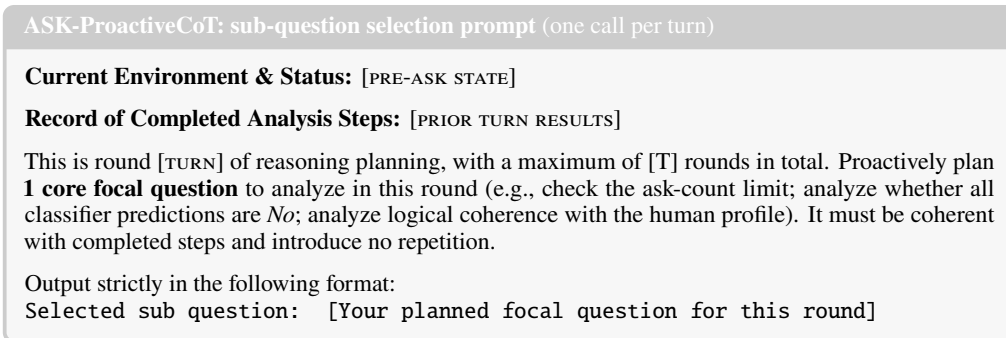

\centering
\begin{minipage}{0.96\linewidth}
\small
\begin{tcolorbox}[
  colback=gray!6, colframe=gray!40,
  title={\textbf{ASK-ProactiveCoT: sub-question selection prompt} (one call per turn)},
  fonttitle=\small, left=4pt, right=4pt, top=4pt, bottom=4pt
]
\textbf{Current Environment \& Status:} [\textsc{pre-ask state}]

\medskip
\textbf{Record of Completed Analysis Steps:} [\textsc{prior turn results}]

\medskip
This is round [\textsc{turn}] of reasoning planning, with a maximum of [\textsc{T}] rounds in total.
Proactively plan \textbf{1 core focal question} to analyze in this round
(e.g., check the ask-count limit; analyze whether all classifier predictions are \textit{No}; analyze logical coherence with the human profile).
It must be coherent with completed steps and introduce no repetition.

\medskip
Output strictly in the following format:

\texttt{Selected sub question: [Your planned focal question for this round]}
\end{tcolorbox}
\end{minipage}
\caption{Sub-question selection prompt for ASK-ProactiveCoT. The companion solve prompt then generates an analysis and interim conclusion for the selected question; a final summarization prompt aggregates all turns into \textsc{Ask} or \textsc{NoAsk}.}
\label{fig:pcot_prompt}
\end{figure}

\paragraph{ASK-ProactiveCoT.}
ASK-ProactiveCoT decomposes the ASK decision into a short sequence of adaptive sub-questions.
At each turn, the model chooses a focal question, such as whether the rejected candidates conflict with the human profile or whether the accepted candidates are sufficient for safe execution.
After the fixed number of turns, a final summarization prompt converts the reasoning record into \textsc{Ask} or \textsc{NoAsk}.
Figure~\ref{fig:pcot_prompt} shows the sub-question selection prompt used at each turn.

\begin{figure}[t]
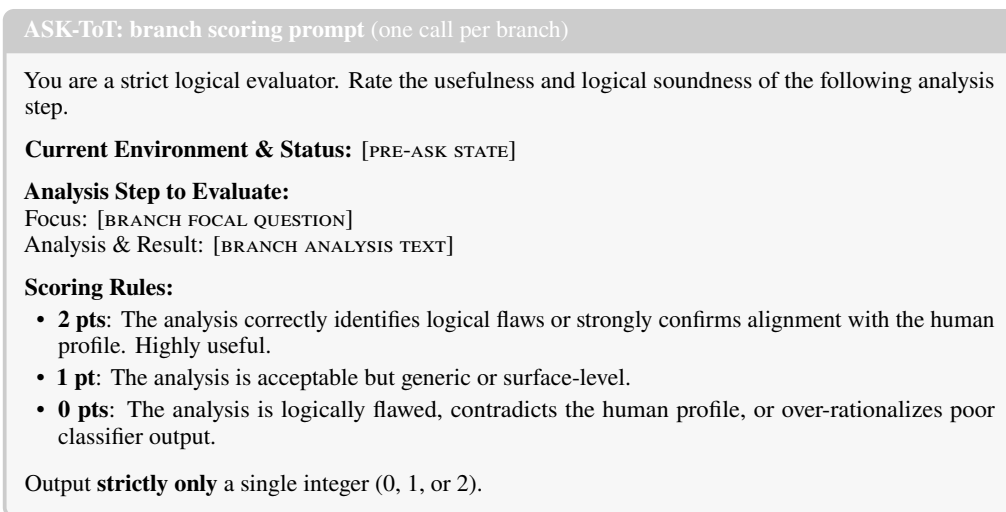

\centering
\begin{minipage}{0.96\linewidth}
\small
\begin{tcolorbox}[
  colback=gray!6, colframe=gray!40,
  title={\textbf{ASK-ToT: branch scoring prompt} (one call per branch)},
  fonttitle=\small, left=4pt, right=4pt, top=4pt, bottom=4pt
]
You are a strict logical evaluator. Rate the usefulness and logical soundness of the following analysis step.

\medskip
\textbf{Current Environment \& Status:} [\textsc{pre-ask state}]

\medskip
\textbf{Analysis Step to Evaluate:}\\
Focus: [\textsc{branch focal question}]\\
Analysis \& Result: [\textsc{branch analysis text}]

\medskip
\textbf{Scoring Rules:}
\begin{itemize}[leftmargin=1.4em, itemsep=0pt, topsep=2pt]
  \item \textbf{2 pts}: The analysis correctly identifies logical flaws or strongly confirms alignment with the human profile. Highly useful.
  \item \textbf{1 pt}: The analysis is acceptable but generic or surface-level.
  \item \textbf{0 pts}: The analysis is logically flawed, contradicts the human profile, or over-rationalizes poor classifier output.
\end{itemize}

\medskip
Output \textbf{strictly only} a single integer (0, 1, or 2).
\end{tcolorbox}
\end{minipage}
\caption{Branch scoring prompt for ASK-ToT. For each of the \(B\) branches at depth \(d\), this prompt is called once to produce the score \(s_{d,b}\in\{0,1,2\}\) used to select the best reasoning path.}
\label{fig:tot_eval_prompt}
\end{figure}

\paragraph{ASK-ToT.}
ASK-ToT performs a small tree search over reasoning branches.
At each depth, it expands multiple candidate analyses, scores their usefulness and logical consistency, and keeps a high-scoring branch for the next depth.
The final decision is generated from the best reasoning path.
This makes the ASK decision less dependent on a single linear chain of thought.
Figure~\ref{fig:tot_eval_prompt} shows the branch-scoring prompt.

\paragraph{ASK-UoT.}
ASK-UoT treats the decision as uncertainty reduction over the candidate set \(\{\textsc{Ask}, \textsc{NoAsk}\}\).
Each turn proposes several verification steps, simulates how each step would reduce the remaining decision uncertainty, executes the most useful step, and updates the candidate set.
If only one candidate decision remains, UoT stops early; otherwise, a final prompt resolves the remaining uncertainty.
Figure~\ref{fig:uot_gen_prompt} shows the verification-step generation prompt.

\begin{figure}[t]
\centering
\begin{minipage}{0.96\linewidth}
\small
\begin{tcolorbox}[
  colback=gray!6, colframe=gray!40,
  title={\textbf{ASK-UoT: verification-step generation prompt} (one call per turn)},
  fonttitle=\small, left=4pt, right=4pt, top=4pt, bottom=4pt
]
\textbf{Current Environment \& Status:} [\textsc{pre-ask state}]

\medskip
\textbf{Completed Verification Records:} [\textsc{prior steps and results}]

\medskip
\textbf{Task:} Generate \textbf{3 different analytical verification steps} to determine whether the robot \textbf{must ask} or \textbf{must not ask} the human.
Each step should be a specific question targeting one of: hard constraints (ask-count limit, time window), trait/profile alignment, or classifier prediction consistency.

\medskip
Output strictly in the following format:

\texttt{Q1: [Verification step 1]}\\
\texttt{Q2: [Verification step 2]}\\
\texttt{Q3: [Verification step 3]}
\end{tcolorbox}
\end{minipage}
\caption{Verification-step generation prompt for ASK-UoT. After this call, each candidate step is simulated to estimate uncertainty reduction; the step with the largest expected reduction is then executed for the final update.}
\label{fig:uot_gen_prompt}
\end{figure}

\subsection{Additional Global Metrics}
\label{app:additional_global_metrics}

\begin{figure*}[p]
    \centering
    \includegraphics[width=\textwidth]{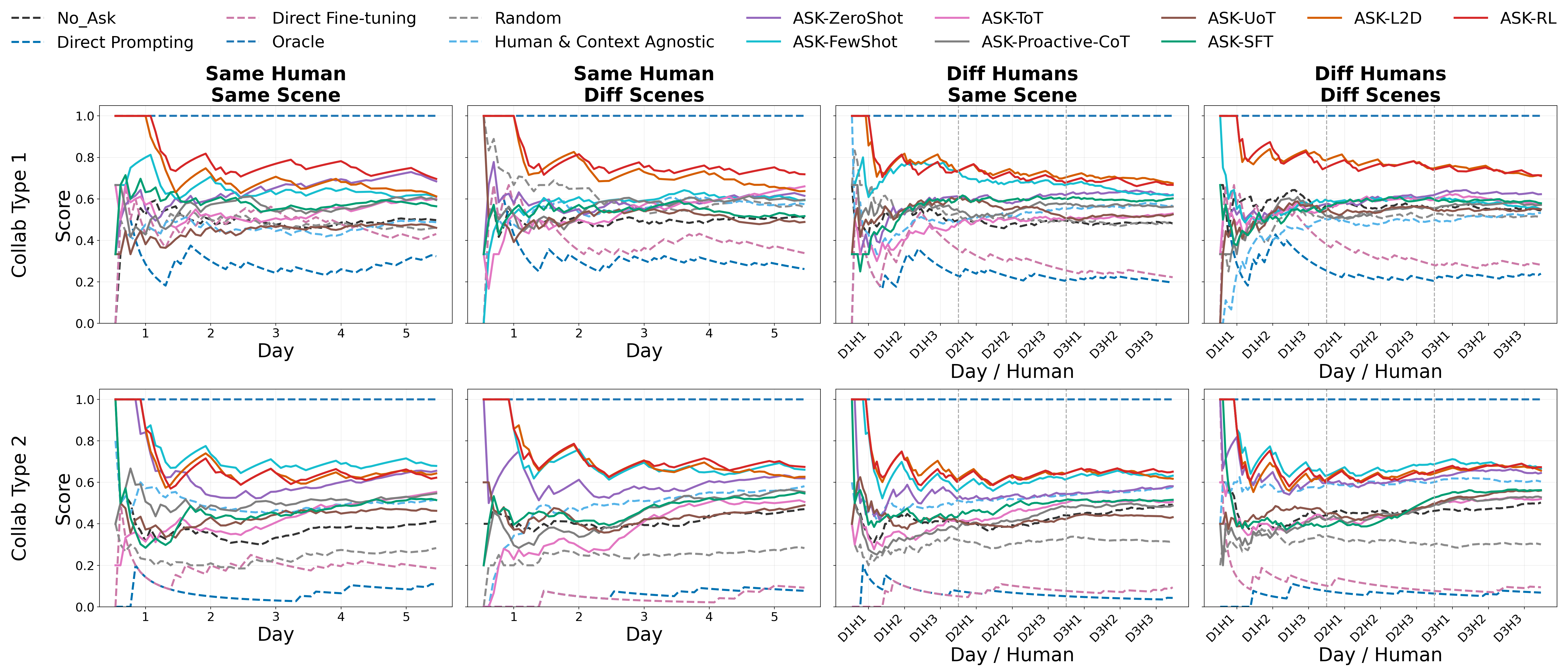}
    \caption{\textbf{Intent-level accuracy across collaboration settings.}
    This figure provides an accuracy-based counterpart to Fig.~\ref{fig:main_intent_f1}. 
    It reports intent prediction accuracy over days for each collaboration type and human-scene setting. 
    Rows correspond to the two collaboration types, and columns correspond to the four human-scene settings. 
    Solid curves denote ASK-based PACT instantiations, while dashed curves denote non-clarification variants.}
    \label{fig:app_main_intent_accuracy}
\end{figure*}

\begin{figure*}[p]
    \centering
    \includegraphics[width=\textwidth]{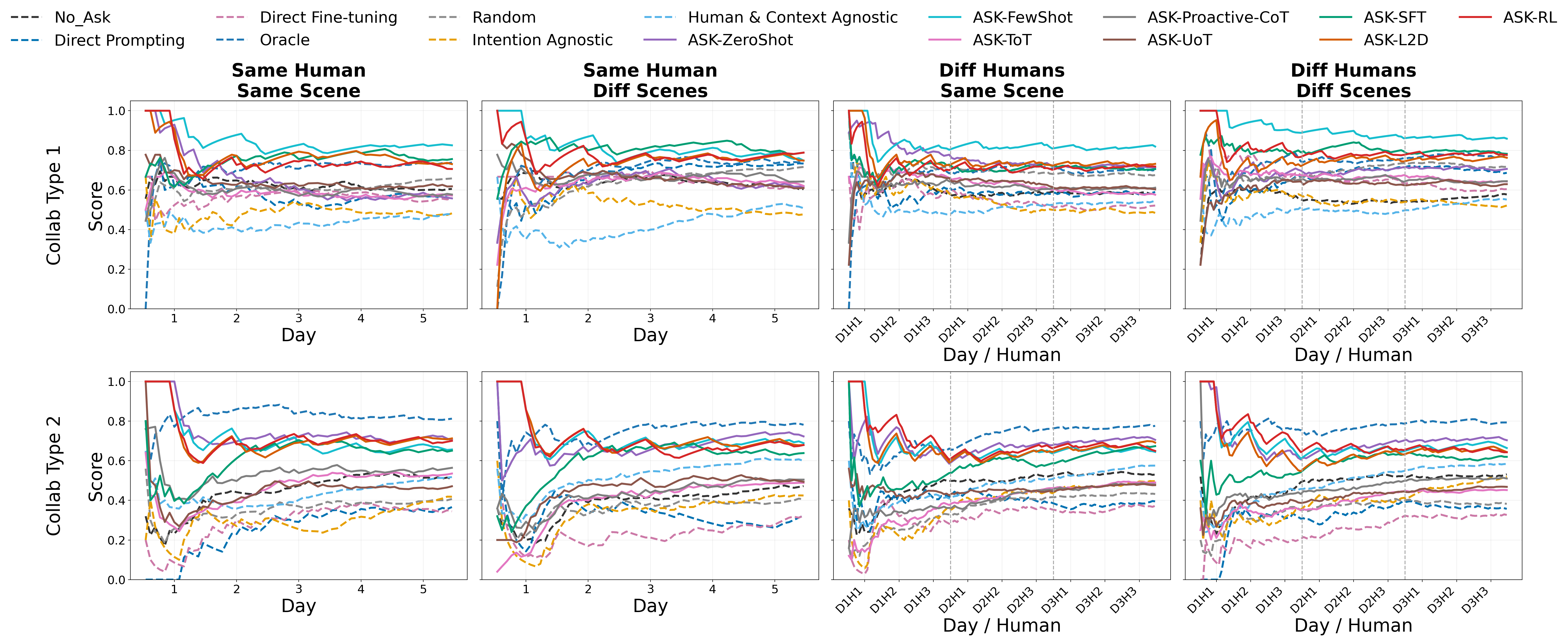}
    \caption{\textbf{Task-level accuracy across collaboration settings.}
    This figure provides an accuracy-based counterpart to Fig.~\ref{fig:main_task_f1}. 
    It reports task-level accuracy over days for each collaboration type and human-scene setting. 
    Rows correspond to the two collaboration types, and columns correspond to the four human-scene settings. 
    Solid curves denote ASK-based PACT instantiations, while dashed curves denote non-clarification variants.}
    \label{fig:app_main_task_accuracy}
\end{figure*}

We provide additional global metrics to complement the main F1-based results in
Figs.~\ref{fig:main_intent_f1} and~\ref{fig:main_task_f1}, as well as the aggregate
clarification-efficiency analysis in Fig.~\ref{fig:efficiency_summary}. 
Figures~\ref{fig:app_main_task_accuracy} and~\ref{fig:app_main_intent_accuracy}
report accuracy-based views of task-level and intent-level performance under the
same collaboration settings. These results follow the same overall pattern as the
main F1 results: ASK-based PACT instantiations are generally more stable than
non-clarification variants, especially when the human partner, the scene, or both
change across days. This indicates that the gains observed in the main figures are
not specific to F1 scoring, but also appear under direct accuracy measurements.

Because the four collaboration settings contain different rollout structures, the
x-axis should be interpreted within each setting. The same-human settings are
reported over five days, while the cross-human settings are reported over
day--human segments. Figure~\ref{fig:app_efficiency_utility_by_setting} further
decomposes the daily Clarification Utility from Fig.~\ref{fig:efficiency_summary}
by collaboration type and setting. This per-setting view shows that clarification
efficiency also varies across days and human-scene configurations, reinforcing the
main conclusion that effective proactive asking depends on selective ask-or-act
decisions rather than simply asking more often.

\subsection{Per-Setting Clarification Efficiency}
\label{app:clarification_by_setting}

We provide per-setting clarification-efficiency analyses to complement the aggregate results in Fig.~\ref{fig:efficiency_summary}. 
The aggregate figure summarizes the overall task accuracy--question trade-off, while the per-setting results below show how this trade-off changes across collaboration types and human-scene settings.

\begin{figure*}[p]
    \centering
    \includegraphics[width=\textwidth]{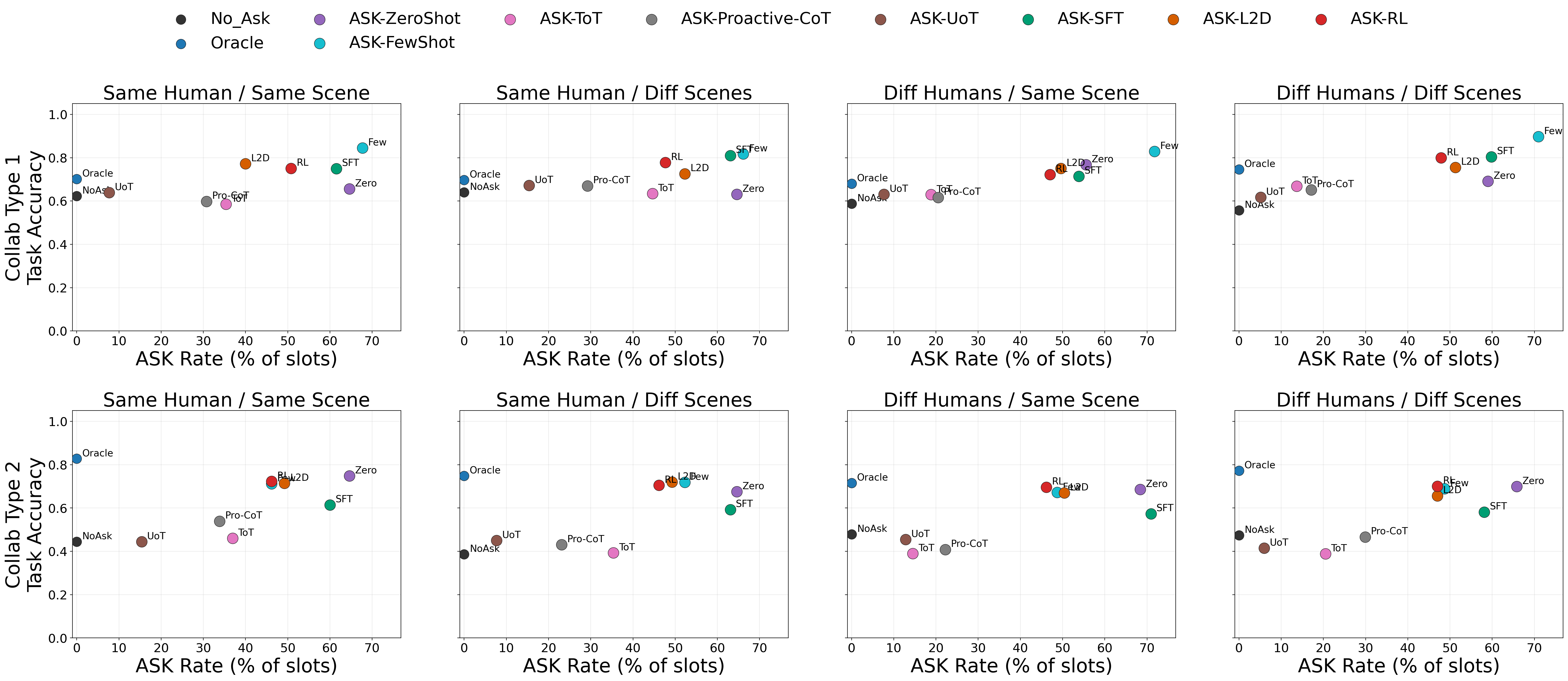}
    \caption{\textbf{Task accuracy versus ASK rate by collaboration setting.}
    Each point corresponds to one method under one collaboration type and human-scene setting. 
    The x-axis reports ASK rate as the percentage of interaction slots with clarification, and the y-axis reports task-level accuracy. 
    Points closer to the upper-left region indicate better clarification efficiency: higher assistance accuracy with fewer questions.}
    \label{fig:accuracy_ask_rate_by_setting}
\end{figure*}

Figure~\ref{fig:accuracy_ask_rate_by_setting} reports the relationship between task accuracy and ASK rate for each collaboration setting. 
Each point corresponds to one method under one collaboration type and human-scene setting. 
The x-axis reports ASK rate as the percentage of interaction slots with clarification, and the y-axis reports task-level accuracy. 
Points closer to the upper-left region indicate better clarification efficiency: higher task accuracy with fewer questions.

The per-setting scatter plots show that higher ASK rate does not necessarily lead to higher task accuracy. 
Some methods ask frequently but do not consistently achieve the best accuracy, suggesting that their clarification decisions are not always well targeted. 
Conversely, very low-ASK methods reduce question frequency but can provide weaker task performance when unresolved intent or task uncertainty remains before acting. 
Across settings, \textbf{ASK-RL} and \textbf{ASK-L2D} tend to occupy more favorable regions of the accuracy--ASK-rate trade-off, achieving strong task accuracy without relying on the highest ASK rates. 
These results support the main conclusion that effective proactive clarification depends on selective ask-or-act decision making rather than simply asking more often.

\begin{figure*}[p]
    \centering
    \includegraphics[width=\textwidth]{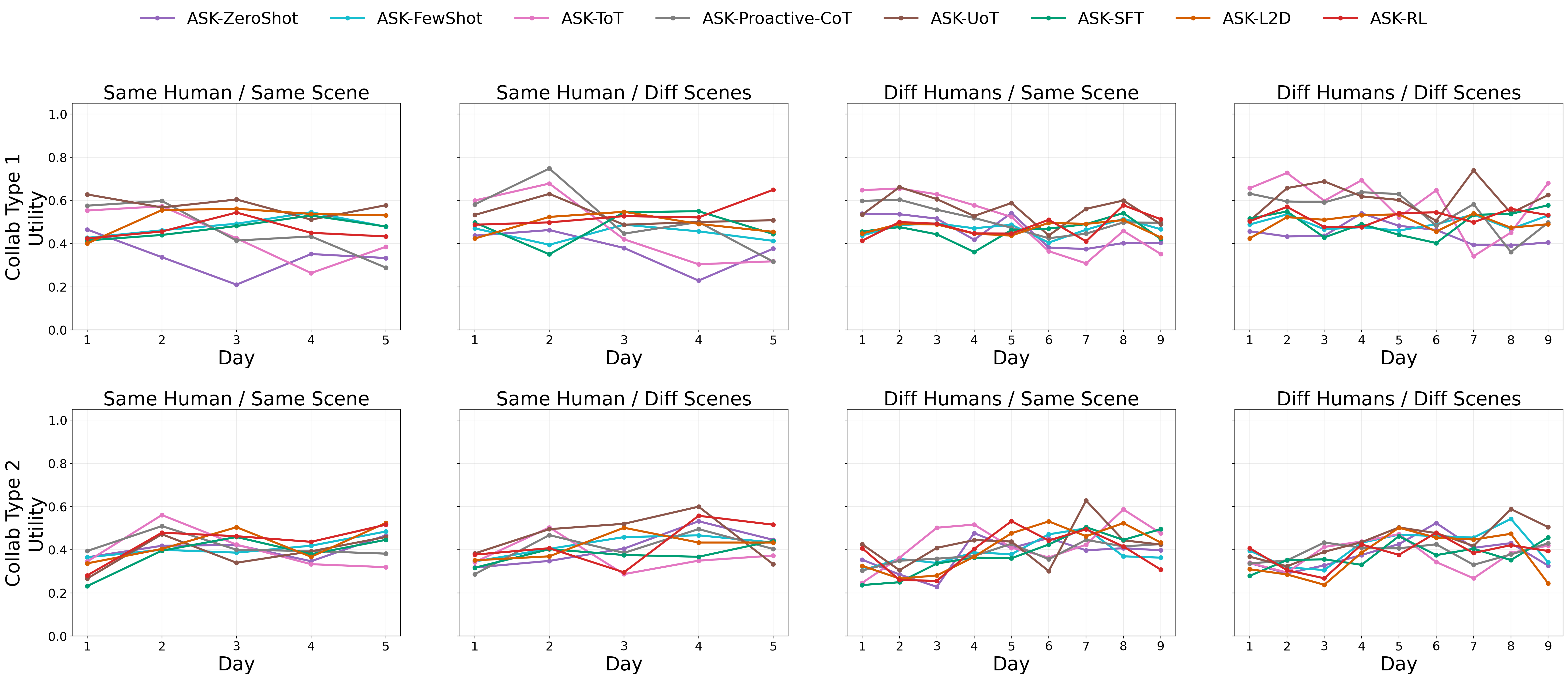}
    \caption{\textbf{Per-setting Clarification Utility across collaboration settings.} 
    Each subplot reports Clarification Utility for ASK-based PACT instantiations under one collaboration type and one human-scene setting. 
    The same-human settings are shown over days, whereas the cross-human settings are shown over day--human segments. 
    Higher values indicate stronger task assistance under a lower clarification burden.}
    \label{fig:app_efficiency_utility_by_setting}
\end{figure*}

Figure~\ref{fig:app_efficiency_utility_by_setting} further reports Clarification Utility for each collaboration type and human-scene setting. 
The same-human settings are shown over days, whereas the cross-human settings are shown over day--human segments. 
The utility trends vary across settings, indicating that clarification efficiency depends on the degree of human-scene variation. 
In more stable same-human settings, several ASK-based methods achieve comparable utility. 
Under cross-human settings, uncertainty-aware prompting such as \textbf{ASK-UoT} often obtains relatively strong utility on some days, suggesting that explicitly reasoning about uncertainty is useful when the assistant cannot rely on a stable human profile. 
At the same time, \textbf{ASK-RL} and \textbf{ASK-L2D} remain comparatively stable across settings, reflecting a more consistent balance between downstream task performance and clarification frequency.

\section{Additional Analysis}
\label{sec:additional_analysis}

\subsection{Per-Setting ASK Usage}
\label{app:per_setting_ask_usage}

\begin{figure*}[p]
    \centering
    \includegraphics[width=\textwidth]{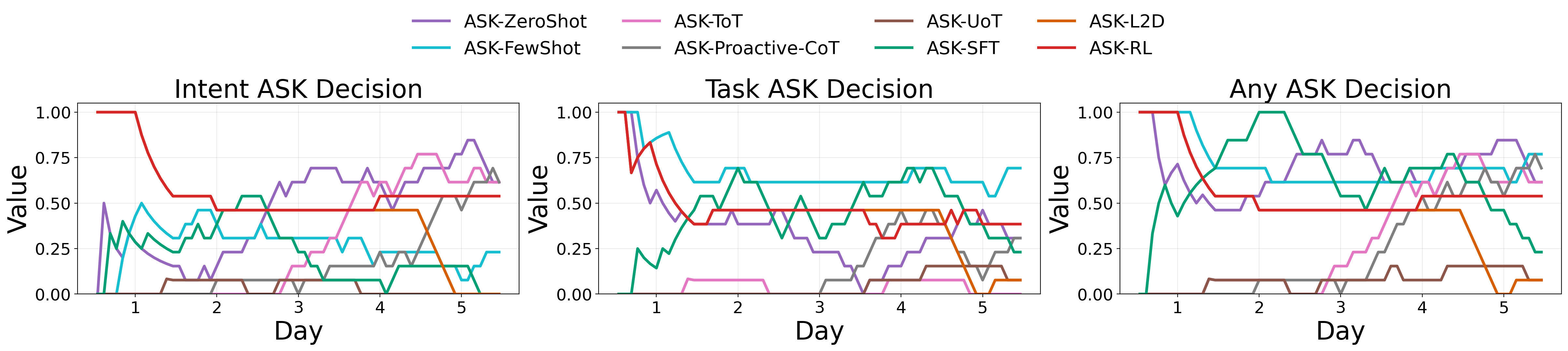}
    \caption{\textbf{ASK usage for collaboration type 1, setting 1.}
    The three panels report intent-level ASK decisions, task-level ASK decisions,
    and overall ASK decisions over the interaction sequence using 13-step moving
    averages.}
    \label{fig:app_ask_usage_c1_s1}
\end{figure*}

\begin{figure*}[p]
    \centering
    \includegraphics[width=\textwidth]{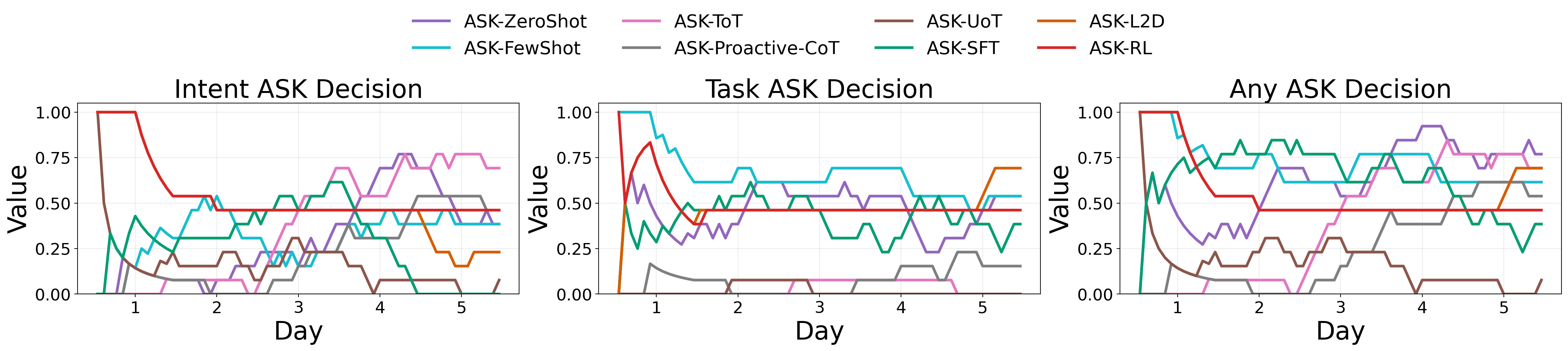}
    \caption{\textbf{ASK usage for collaboration type 1, setting 2.}
    The three panels report intent-level ASK decisions, task-level ASK decisions,
    and overall ASK decisions over the interaction sequence using 13-step moving
    averages.}
    \label{fig:app_ask_usage_c1_s2}
\end{figure*}

\begin{figure*}[p]
    \centering
    \includegraphics[width=\textwidth]{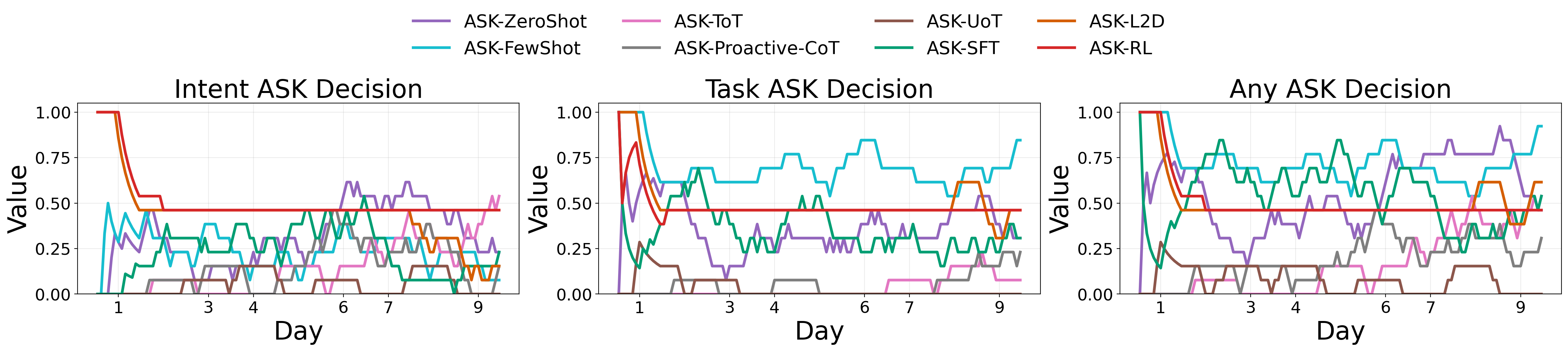}
    \caption{\textbf{ASK usage for collaboration type 1, setting 3.}
    The three panels report intent-level ASK decisions, task-level ASK decisions,
    and overall ASK decisions over the interaction sequence using 13-step moving
    averages.}
    \label{fig:app_ask_usage_c1_s3}
\end{figure*}

\begin{figure*}[p]
    \centering
    \includegraphics[width=\textwidth]{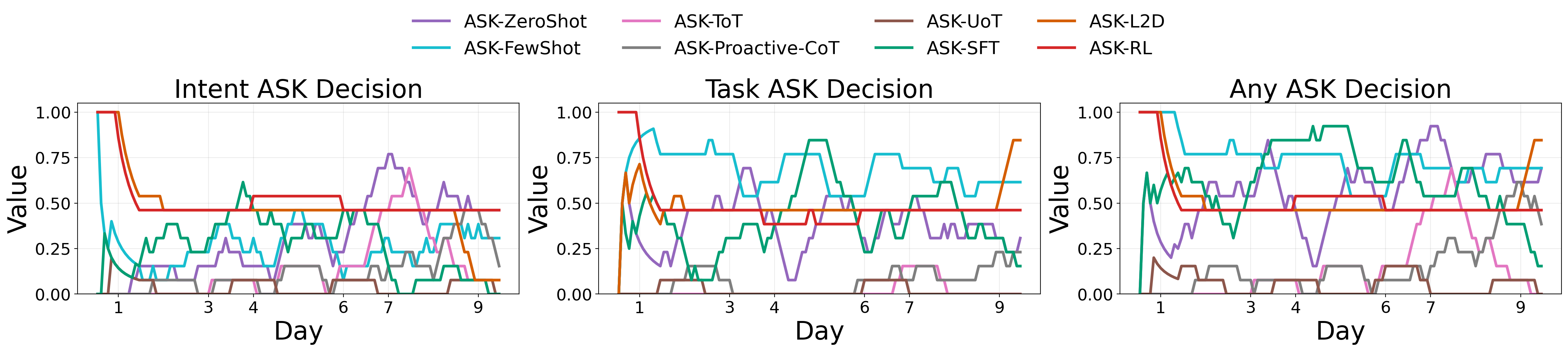}
    \caption{\textbf{ASK usage for collaboration type 1, setting 4.}
    The three panels report intent-level ASK decisions, task-level ASK decisions,
    and overall ASK decisions over the interaction sequence using 13-step moving
    averages.}
    \label{fig:app_ask_usage_c1_s4}
\end{figure*}

\begin{figure*}[p]
    \centering
    \includegraphics[width=\textwidth]{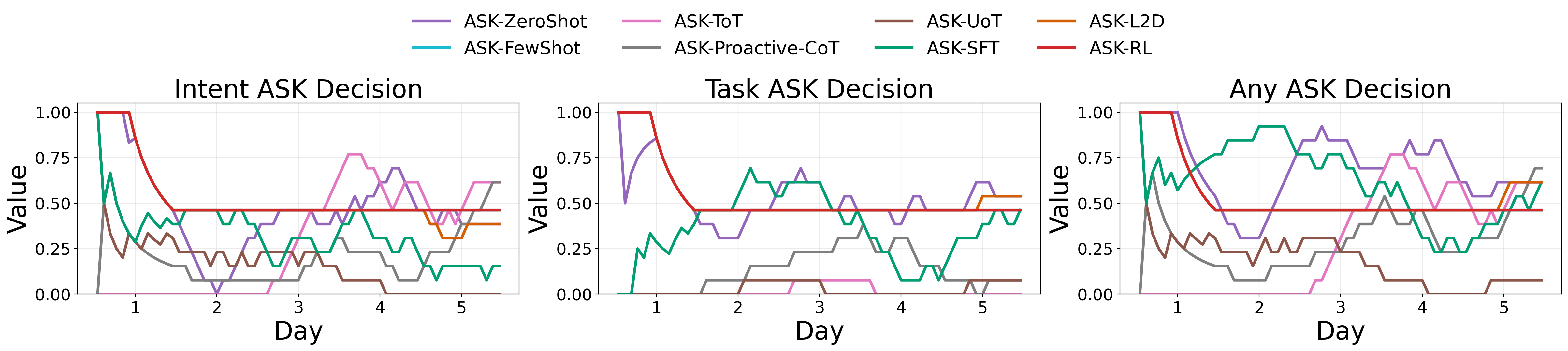}
    \caption{\textbf{ASK usage for collaboration type 2, setting 1.}
    The three panels report intent-level ASK decisions, task-level ASK decisions,
    and overall ASK decisions over the interaction sequence using 13-step moving
    averages.}
    \label{fig:app_ask_usage_c2_s1}
\end{figure*}

\begin{figure*}[p]
    \centering
    \includegraphics[width=\textwidth]{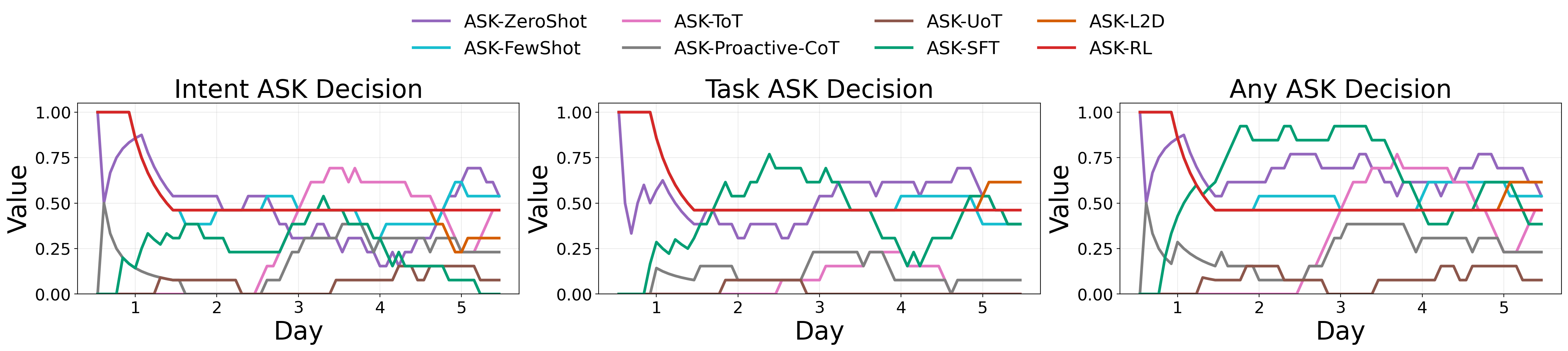}
    \caption{\textbf{ASK usage for collaboration type 2, setting 2.}
    The three panels report intent-level ASK decisions, task-level ASK decisions,
    and overall ASK decisions over the interaction sequence using 13-step moving
    averages.}
    \label{fig:app_ask_usage_c2_s2}
\end{figure*}

\begin{figure*}[p]
    \centering
    \includegraphics[width=\textwidth]{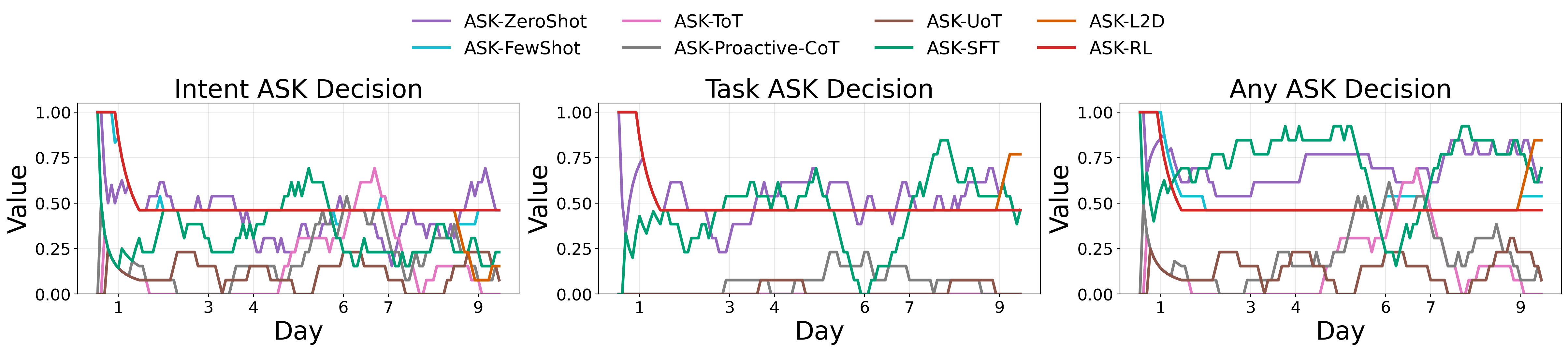}
    \caption{\textbf{ASK usage for collaboration type 2, setting 3.}
    The three panels report intent-level ASK decisions, task-level ASK decisions,
    and overall ASK decisions over the interaction sequence using 13-step moving
    averages.}
    \label{fig:app_ask_usage_c2_s3}
\end{figure*}

\begin{figure*}[p]
    \centering
    \includegraphics[width=\textwidth]{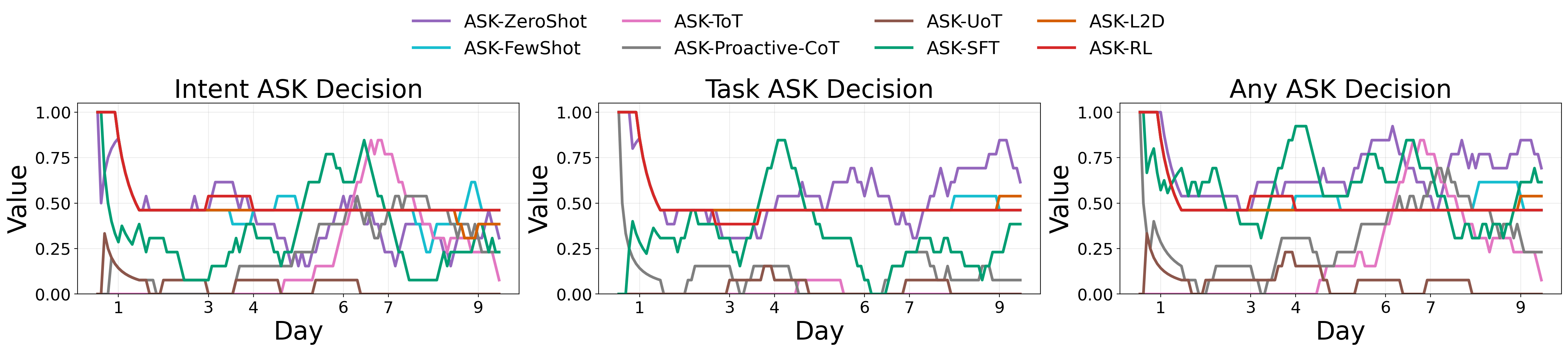}
    \caption{\textbf{ASK usage for collaboration type 2, setting 4.}
    The three panels report intent-level ASK decisions, task-level ASK decisions,
    and overall ASK decisions over the interaction sequence using 13-step moving
    averages.}
    \label{fig:app_ask_usage_c2_s4}
\end{figure*}

We first analyze how ASK-based methods allocate clarification decisions over the interaction sequence. 
Figures~\ref{fig:app_ask_usage_c1_s1}--\ref{fig:app_ask_usage_c2_s4} report 13-step moving-average trends of intent-level ASK decisions, task-level ASK decisions, and overall ASK decisions for each collaboration type and setting. 
Different from the aggregate ASK rate in Fig.~\ref{fig:efficiency_summary}, these plots show whether clarification is mainly used to resolve intent uncertainty, task-level uncertainty, or both.

Across settings, the methods show distinct allocation patterns. 
ASK-FewShot often maintains high task-level and overall ASK usage, especially when the scene or human partner changes, suggesting that few-shot prompting tends to treat task ambiguity as persistent. 
ASK-ZeroShot and ASK-ToT show more delayed or fluctuating ASK behavior, with ASK rates that are often low in early segments but increase later. 
ASK-SFT also fluctuates substantially, indicating that supervised ASK prediction can capture useful signals but is sensitive to changes in the human-scene configuration.

ASK-UoT and ASK-Proactive-CoT are generally more conservative: their intent-level and task-level ASK rates often remain low for long segments, which reduces question frequency but may leave unresolved uncertainty in harder settings. 
In contrast, ASK-RL and ASK-L2D often begin with higher ASK usage and then reduce or stabilize their question frequency as the sequence progresses. 
This pattern suggests that learned ask-or-act policies use early clarification to reduce uncertainty and then rely more on accumulated interaction history.

These results explain why high ASK rate alone does not guarantee better clarification efficiency. 
Excessive task-level asking can increase the clarification burden without consistently improving assistance, while overly conservative asking can leave the assistant acting under unresolved intent or task uncertainty.

\subsection{Per-Setting ASK Policy Quality}
\label{app:per_setting_ask_policy_quality}

\begin{figure*}[p]
    \centering
    \includegraphics[width=\textwidth]{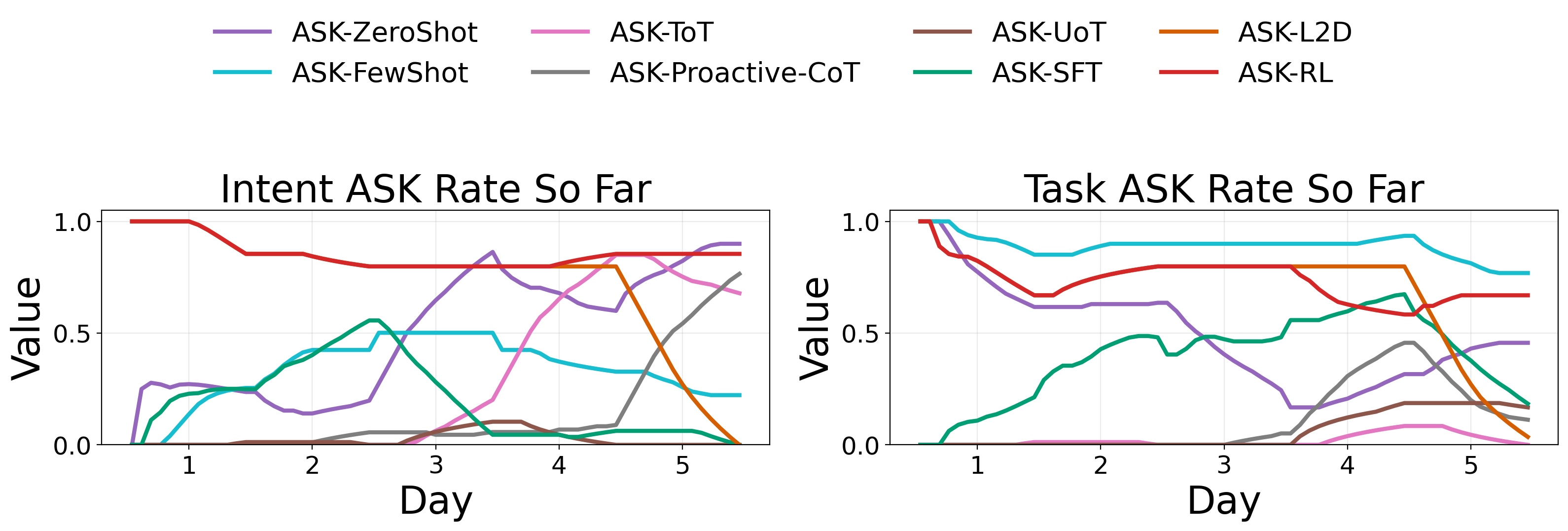}
    \caption{\textbf{ASK policy quality for collaboration type 1, setting 1.}
    The two panels report cumulative intent-level and task-level ASK rates over
    the interaction sequence.}
    \label{fig:app_ask_policy_quality_c1_s1}
\end{figure*}

\begin{figure*}[p]
    \centering
    \includegraphics[width=\textwidth]{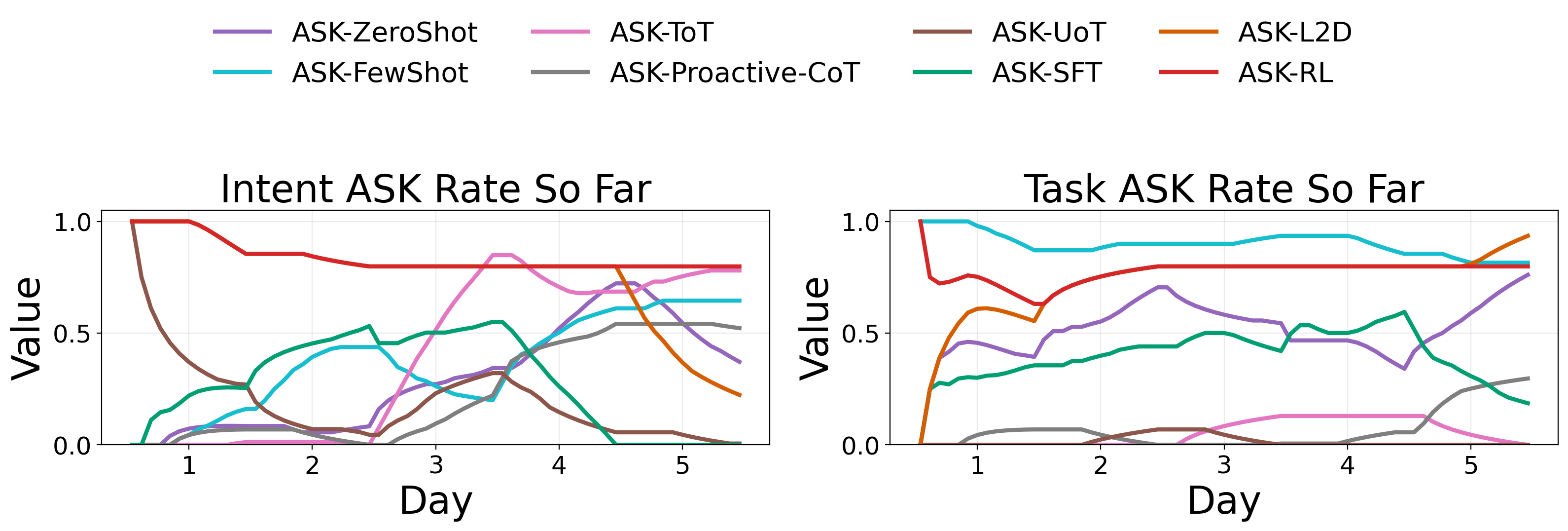}
    \caption{\textbf{ASK policy quality for collaboration type 1, setting 2.}
    The two panels report cumulative intent-level and task-level ASK rates over
    the interaction sequence.}
    \label{fig:app_ask_policy_quality_c1_s2}
\end{figure*}

\begin{figure*}[p]
    \centering
    \includegraphics[width=\textwidth]{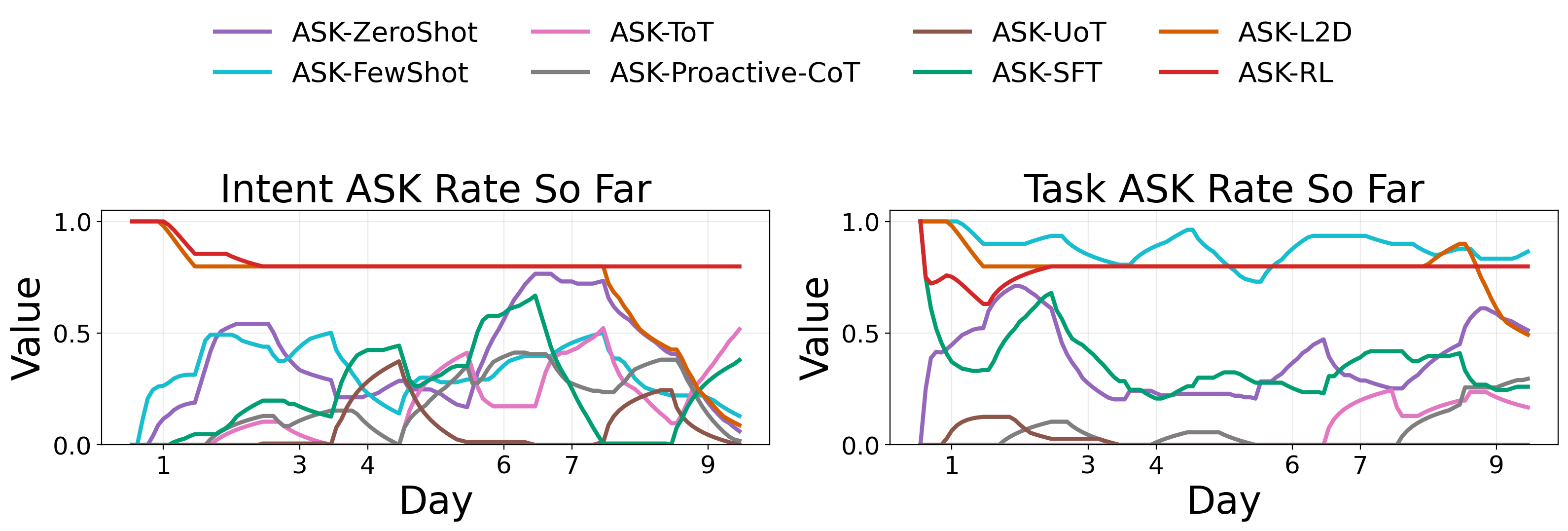}
    \caption{\textbf{ASK policy quality for collaboration type 1, setting 3.}
    The two panels report cumulative intent-level and task-level ASK rates over
    the interaction sequence.}
    \label{fig:app_ask_policy_quality_c1_s3}
\end{figure*}

\begin{figure*}[p]
    \centering
    \includegraphics[width=\textwidth]{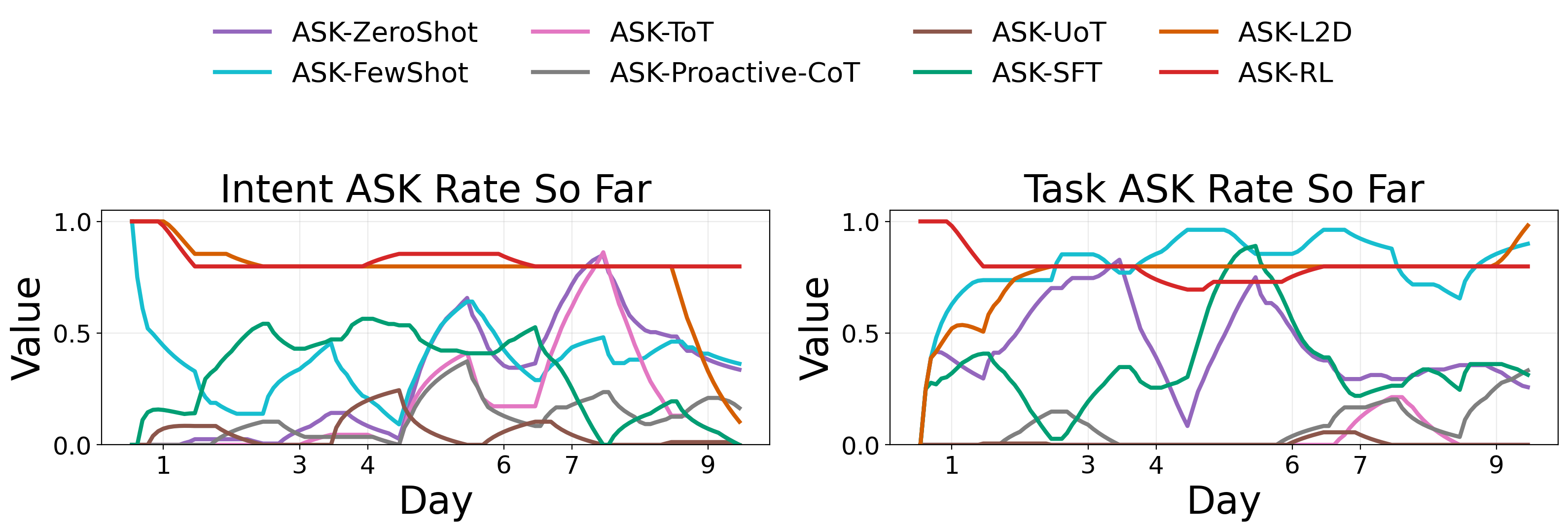}
    \caption{\textbf{ASK policy quality for collaboration type 1, setting 4.}
    The two panels report cumulative intent-level and task-level ASK rates over
    the interaction sequence.}
    \label{fig:app_ask_policy_quality_c1_s4}
\end{figure*}

\begin{figure*}[p]
    \centering
    \includegraphics[width=\textwidth]{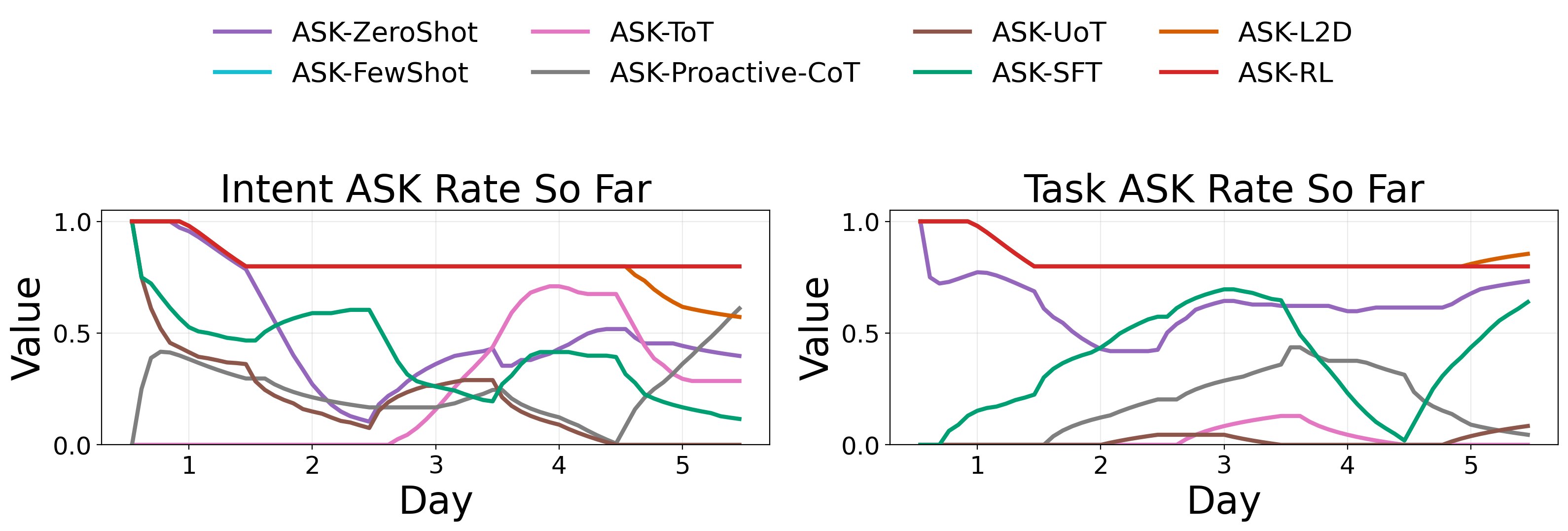}
    \caption{\textbf{ASK policy quality for collaboration type 2, setting 1.}
    The two panels report cumulative intent-level and task-level ASK rates over
    the interaction sequence.}
    \label{fig:app_ask_policy_quality_c2_s1}
\end{figure*}

\begin{figure*}[p]
    \centering
    \includegraphics[width=\textwidth]{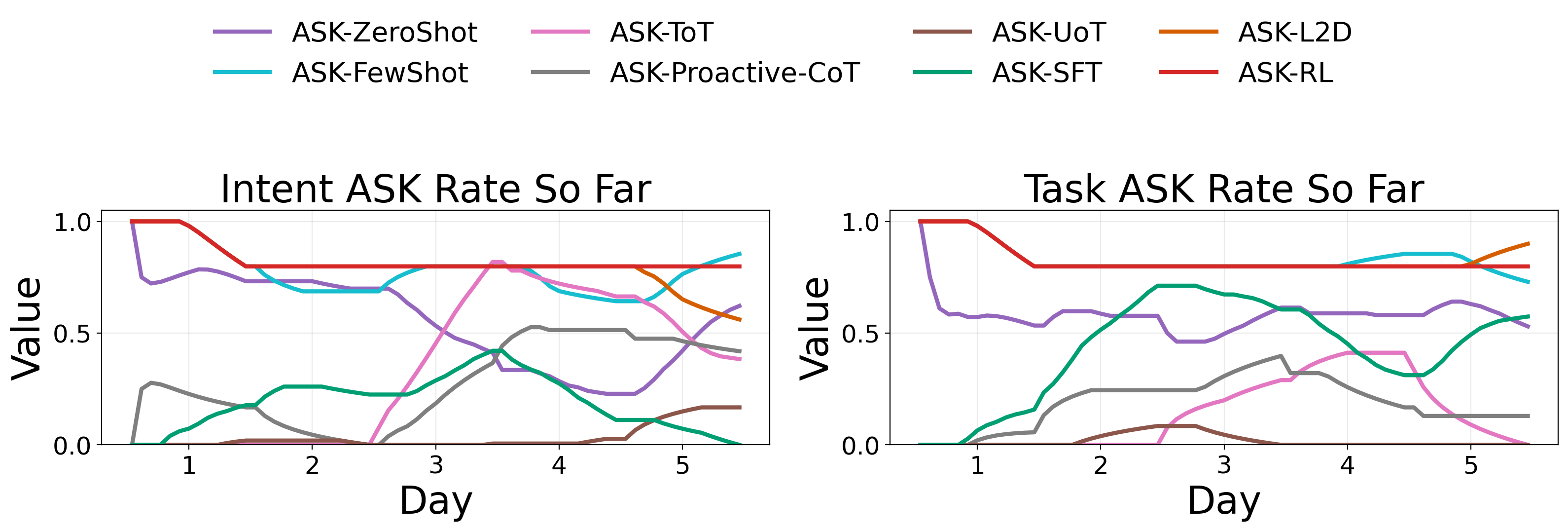}
    \caption{\textbf{ASK policy quality for collaboration type 2, setting 2.}
    The two panels report cumulative intent-level and task-level ASK rates over
    the interaction sequence.}
    \label{fig:app_ask_policy_quality_c2_s2}
\end{figure*}

\begin{figure*}[p]
    \centering
    \includegraphics[width=\textwidth]{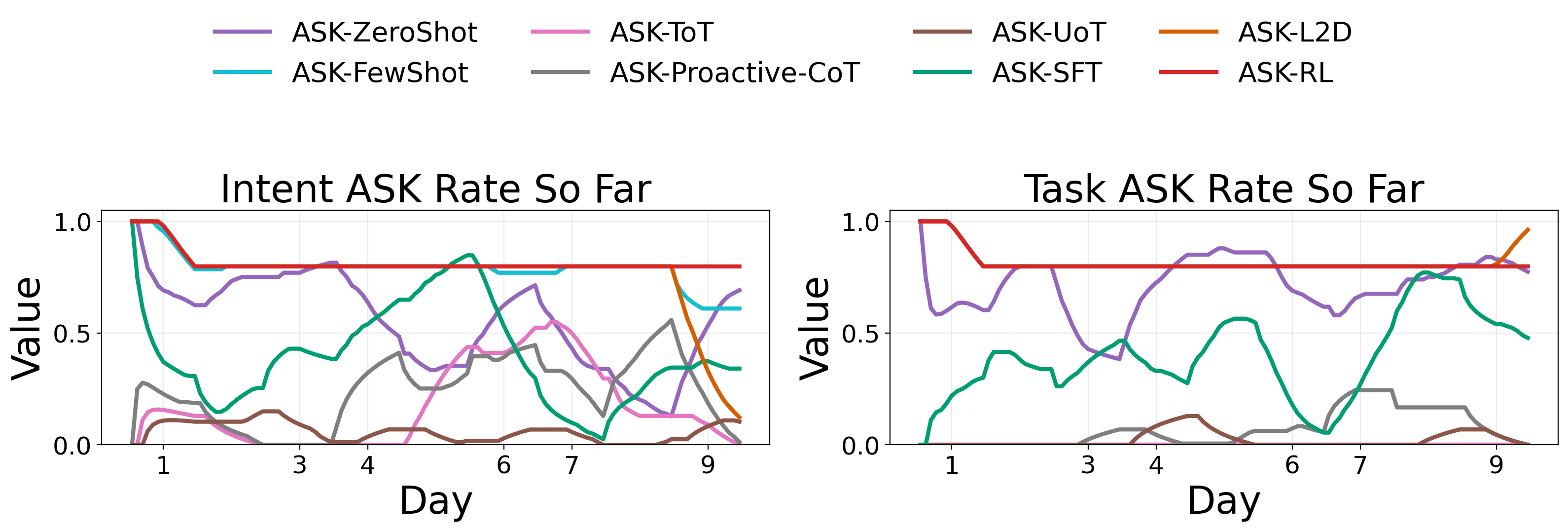}
    \caption{\textbf{ASK policy quality for collaboration type 2, setting 3.}
    The two panels report cumulative intent-level and task-level ASK rates over
    the interaction sequence.}
    \label{fig:app_ask_policy_quality_c2_s3}
\end{figure*}

\begin{figure*}[p]
    \centering
    \includegraphics[width=\textwidth]{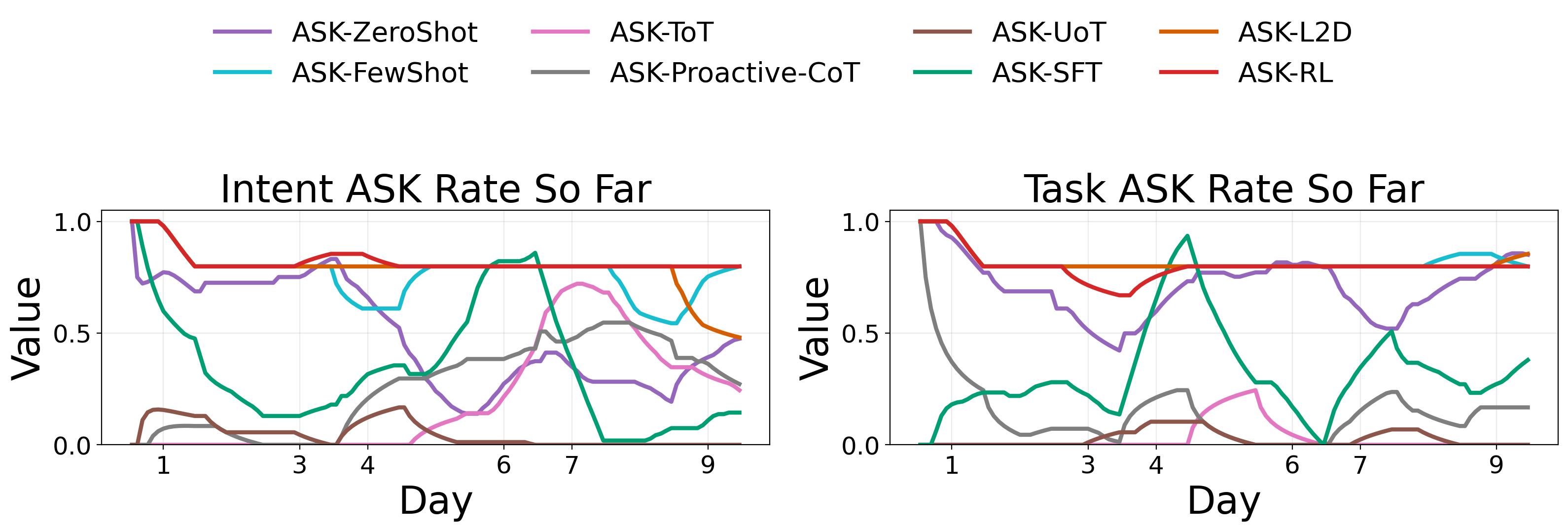}
    \caption{\textbf{ASK policy quality for collaboration type 2, setting 4.}
    The two panels report cumulative intent-level and task-level ASK rates over
    the interaction sequence.}
    \label{fig:app_ask_policy_quality_c2_s4}
\end{figure*}

We next examine the cumulative behavior of ASK policies in each collaboration setting. 
Different from Appendix~\ref{app:per_setting_ask_usage}, which reports local moving-average ASK trends, Figs.~\ref{fig:app_ask_policy_quality_c1_s1}--\ref{fig:app_ask_policy_quality_c2_s4} report cumulative intent-level and task-level ASK rates over the interaction sequence. 
These plots show whether a method gradually reduces, maintains, or increases its tendency to ask as interaction history accumulates.

The cumulative profiles reveal clear differences among ASK policies. 
ASK-FewShot often maintains high task-level ASK rates, particularly under scene or human variation, suggesting a persistent tendency to ask about task or predicate uncertainty. 
ASK-ZeroShot and ASK-ToT show less stable cumulative trajectories, with ASK rates that can remain low early and increase later. 
ASK-SFT also varies substantially across settings, suggesting that supervised ASK prediction is sensitive to the current human-scene configuration.

Policy-based methods show more structured cumulative behavior. 
ASK-RL and ASK-L2D often start with high ASK rates and then stabilize or reduce cumulative question frequency. 
This indicates that early clarification can help reduce uncertainty before later decisions rely more on accumulated context. 
By contrast, ASK-UoT and ASK-Proactive-CoT often keep low cumulative ASK rates for long segments, which reduces interruptions but can lead to under-asking when uncertainty remains unresolved.

Together with the main performance curves and Clarification Utility results, these figures show that ASK policy quality is not determined by asking more often. 
A stronger policy should calibrate its ASK rate over time according to interaction history and collaboration difficulty.

\subsection{Per-Setting ASK Impact}
\label{app:per_setting_ask_impact}

\begin{figure*}[p]
    \centering
    \includegraphics[width=\textwidth]{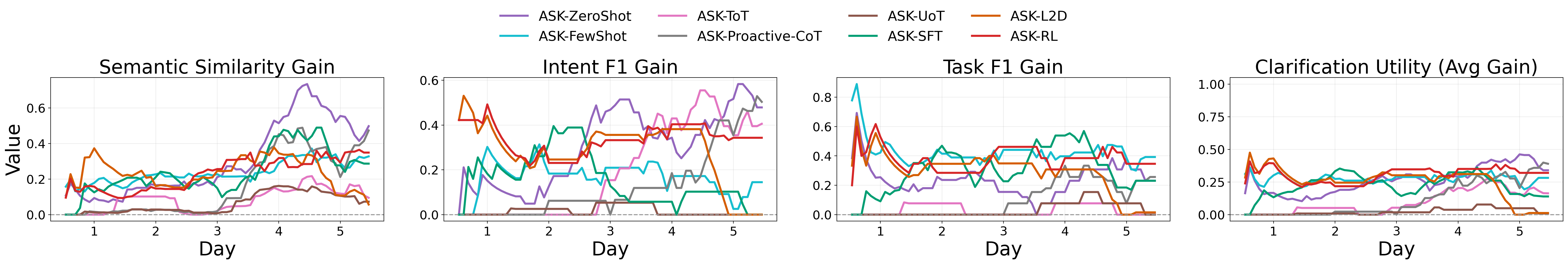}
    \caption{\textbf{ASK impact for collaboration type 1, setting 1.}
    The four panels report semantic similarity gain, intent F1 gain, task F1
    gain, and average Clarification Utility gain after ASK decisions.}
    \label{fig:app_ask_impact_c1_s1}
\end{figure*}

\begin{figure*}[p]
    \centering
    \includegraphics[width=\textwidth]{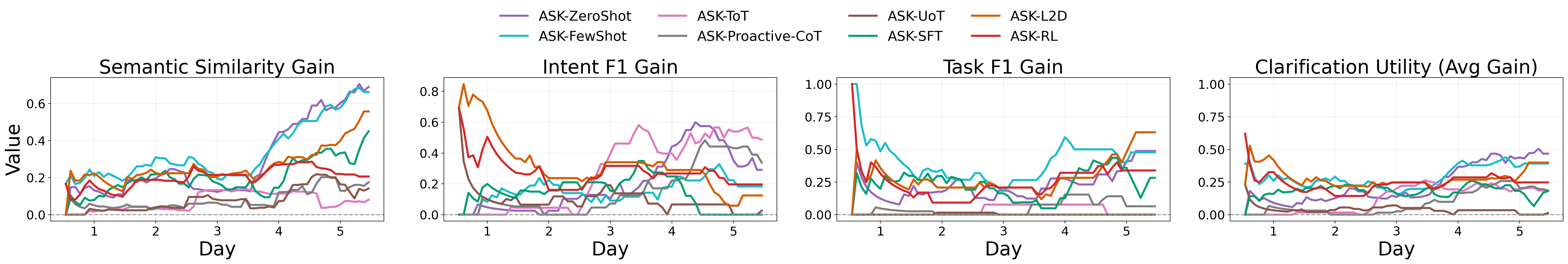}
    \caption{\textbf{ASK impact for collaboration type 1, setting 2.}
    The four panels report semantic similarity gain, intent F1 gain, task F1
    gain, and average Clarification Utility gain after ASK decisions.}
    \label{fig:app_ask_impact_c1_s2}
\end{figure*}

\begin{figure*}[p]
    \centering
    \includegraphics[width=\textwidth]{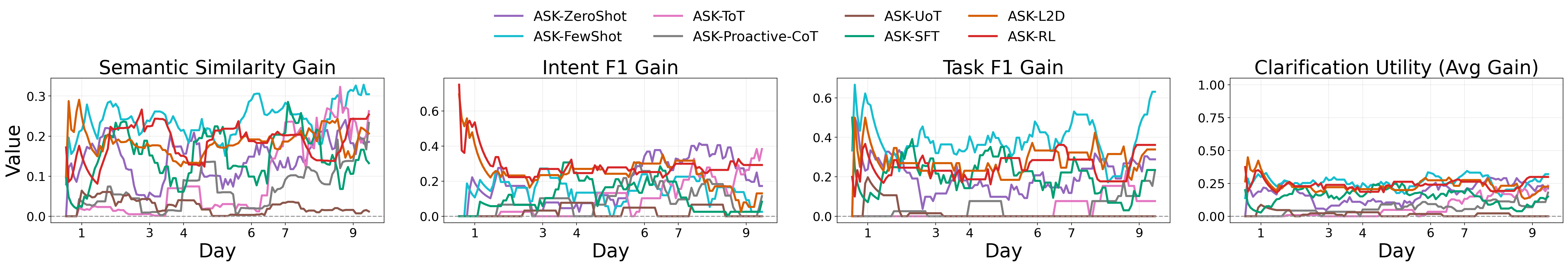}
    \caption{\textbf{ASK impact for collaboration type 1, setting 3.}
    The four panels report semantic similarity gain, intent F1 gain, task F1
    gain, and average Clarification Utility gain after ASK decisions.}
    \label{fig:app_ask_impact_c1_s3}
\end{figure*}

\begin{figure*}[p]
    \centering
    \includegraphics[width=\textwidth]{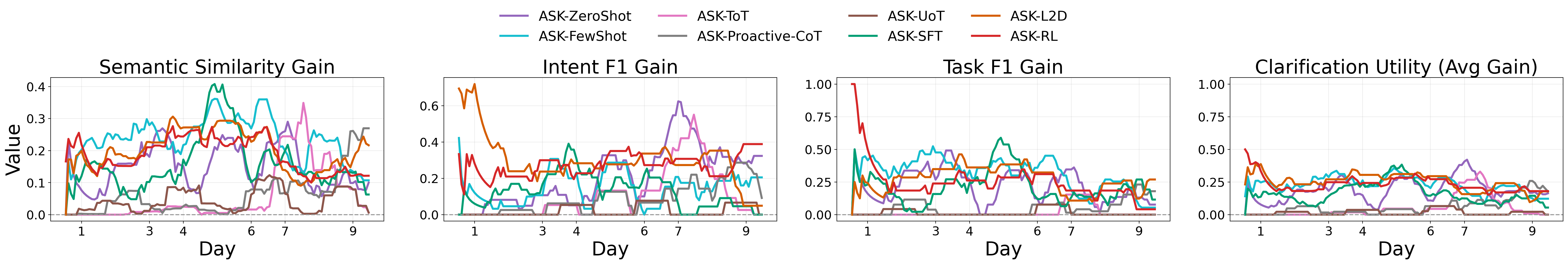}
    \caption{\textbf{ASK impact for collaboration type 1, setting 4.}
    The four panels report semantic similarity gain, intent F1 gain, task F1
    gain, and average Clarification Utility gain after ASK decisions.}
    \label{fig:app_ask_impact_c1_s4}
\end{figure*}

\begin{figure*}[p]
    \centering
    \includegraphics[width=\textwidth]{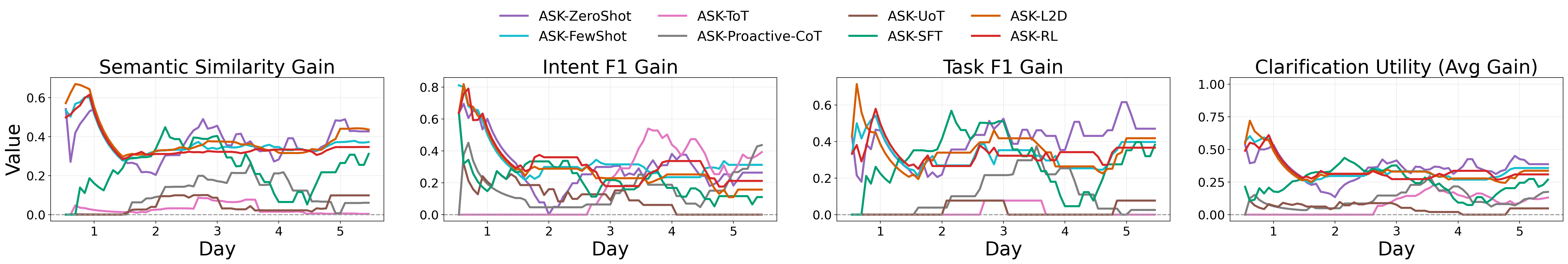}
    \caption{\textbf{ASK impact for collaboration type 2, setting 1.}
    The four panels report semantic similarity gain, intent F1 gain, task F1
    gain, and average Clarification Utility gain after ASK decisions.}
    \label{fig:app_ask_impact_c2_s1}
\end{figure*}

\begin{figure*}[p]
    \centering
    \includegraphics[width=\textwidth]{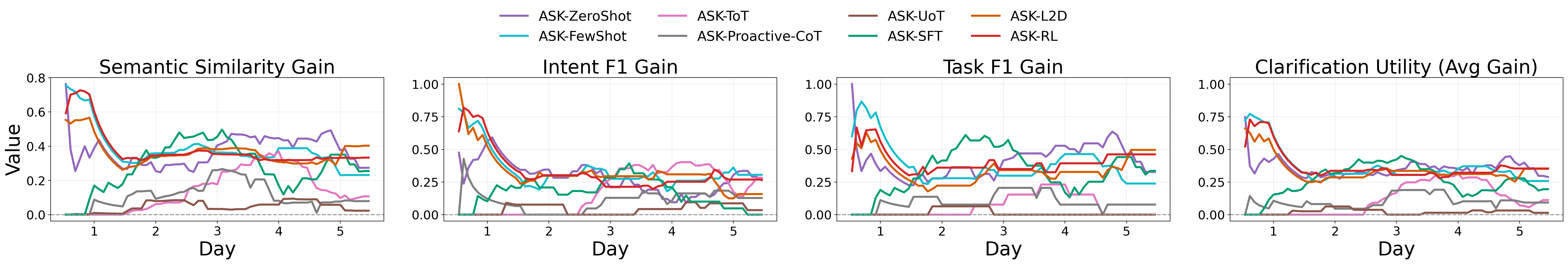}
    \caption{\textbf{ASK impact for collaboration type 2, setting 2.}
    The four panels report semantic similarity gain, intent F1 gain, task F1
    gain, and average Clarification Utility gain after ASK decisions.}
    \label{fig:app_ask_impact_c2_s2}
\end{figure*}

\begin{figure*}[p]
    \centering
    \includegraphics[width=\textwidth]{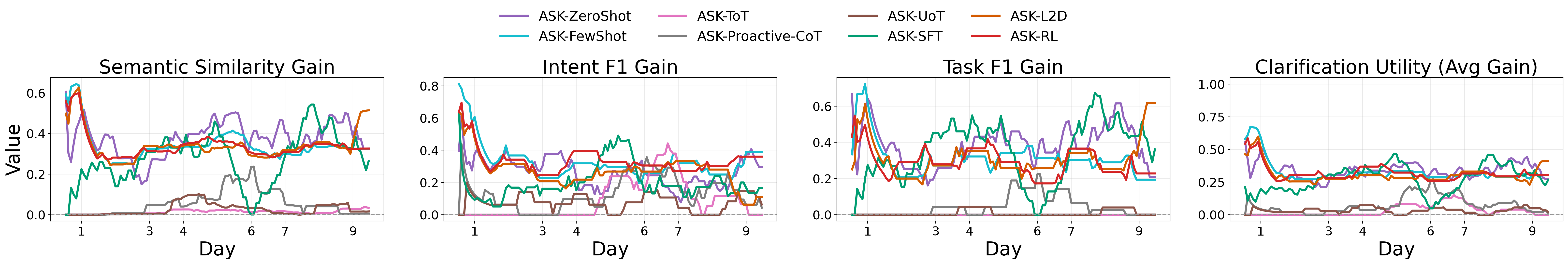}
    \caption{\textbf{ASK impact for collaboration type 2, setting 3.}
    The four panels report semantic similarity gain, intent F1 gain, task F1
    gain, and average Clarification Utility gain after ASK decisions.}
    \label{fig:app_ask_impact_c2_s3}
\end{figure*}

\begin{figure*}[p]
    \centering
    \includegraphics[width=\textwidth]{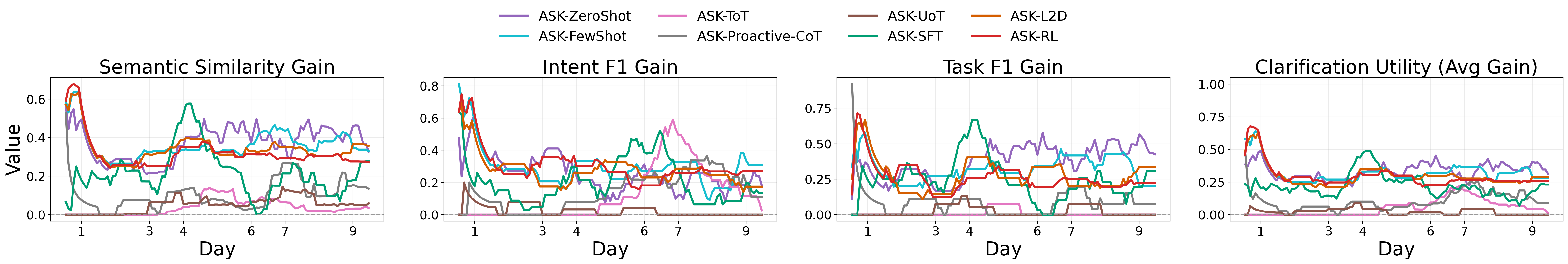}
    \caption{\textbf{ASK impact for collaboration type 2, setting 4.}
    The four panels report semantic similarity gain, intent F1 gain, task F1
    gain, and average Clarification Utility gain after ASK decisions.}
    \label{fig:app_ask_impact_c2_s4}
\end{figure*}

We finally evaluate whether ASK decisions produce measurable gains after clarification. 
Figures~\ref{fig:app_ask_impact_c1_s1}--\ref{fig:app_ask_impact_c2_s4} report per-setting ASK impact from four perspectives: semantic similarity gain, intent F1 gain, task F1 gain, and average Clarification Utility gain. 
Compared with ASK usage and cumulative ASK-rate plots, these figures provide a more direct diagnostic of whether clarification is beneficial rather than merely frequent.

Across settings, frequent asking does not always produce stable post-ASK gains. 
ASK-FewShot often obtains relatively large semantic similarity and task-level gains, indicating that frequent task-level questions can recover useful missing information. 
However, its gains also fluctuate across settings, showing that high ASK usage does not always translate into stable utility improvement. 
ASK-ZeroShot and ASK-ToT show more variable impact patterns, with gains appearing in some segments but less consistently across intent-level, task-level, and utility metrics.

Training-based and policy-based methods show more informative trends. 
ASK-SFT can produce strong task-level gains in some settings, but its impact varies substantially over time. 
ASK-RL and ASK-L2D often show larger early gains and then settle into moderate positive improvements, consistent with their cumulative ASK behavior: early clarification reduces uncertainty, while later decisions rely more on accumulated interaction history. 
In contrast, ASK-UoT and ASK-Proactive-CoT frequently remain near low-gain regions, suggesting that conservative asking can fail to obtain enough useful clarification when intent or task uncertainty remains unresolved.

These results reinforce the main conclusion that effective proactive asking requires useful and selective clarification. 
A strong ASK policy should not only control the number of questions, but also ensure that questions improve semantic alignment, intent prediction, task prediction, and overall clarification utility.

\end{document}